\documentclass[sigconf,screen,authorversion,nonacm]{acmart}
\usepackage{multirow}
\usepackage{framed}
\usepackage{algorithm}
\usepackage{algorithmicx}
\usepackage{algpseudocode}
\usepackage[normalem]{ulem}
\useunder{\uline}{\ul}{}
\usepackage{array}
\usepackage{tabularx}
\algrenewcommand\algorithmicindent{0.5em}
\newcommand{\dataset}[1]{\texttt{{LiteraryTaste}}}
\makeatletter
\newcommand{\adaptivetablesize}{%
  \if@ACM@manuscript\footnotesize
  \else\small\fi
}
\makeatother

\def\thickhline{\noalign{\hrule height1pt}}
\AtBeginDocument{%
  }

\setcopyright{acmlicensed}
\copyrightyear{2018}
\acmYear{2018}
\acmDOI{XXXXXXX.XXXXXXX}
\acmConference[Conference acronym 'XX]{Make sure to enter the correct
  conference title from your rights confirmation email}{June 03--05,
  2018}{Woodstock, NY}
\acmISBN{978-1-4503-XXXX-X/2018/06}

\begin{document}

\title{\dataset{}: A Preference Dataset for Creative Writing Personalization}


\author{John Joon Young Chung}
\affiliation{%
  \institution{Midjourney}
  \city{San Francisco}
    \state{CA}
  \country{USA}}
\email{jchung@midjourney.org}

\author{Vishakh Padmakumar}
\affiliation{%
  \institution{Stanford University}
  \city{Palo Alto}
    \state{CA}
  \country{USA}}
\email{vishakhp@stanford.edu}

\author{Melissa Roemmele}
\affiliation{%
  \institution{Midjourney}
  \city{San Francisco}
    \state{CA}
  \country{USA}}
\email{mroemmele@midjourney.org}

\author{Yi Wang}
\affiliation{%
  \institution{Midjourney}
  \city{San Francisco}
    \state{CA}
  \country{USA}}
\email{ywang@midjourney.org}

\author{Yuqian Sun}
\affiliation{%
  \institution{Midjourney}
  \city{San Francisco}
    \state{CA}
  \country{USA}}
\email{ysun@midjourney.org}

\author{Tiffany Wang}
\affiliation{%
  \institution{Midjourney}
  \city{San Francisco}
    \state{CA}
  \country{USA}}
\email{twang@midjourney.org}

\author{Shm Garanganao Almeda}
\affiliation{
    \institution{UC Berkeley}
    \city{Berkeley}
    \state{CA}
    \country{USA}
}
\email{shm.almeda@berkeley.edu}

\author{Brett A. Halperin}
\affiliation{
    \institution{University of Washington}
    \city{Seattle}
    \state{WA}
    \country{USA}
}
\email{bhalp@uw.edu}

\author{Yuwen Lu}
\affiliation{
    \institution{University of Notre Dame}
    \city{Notre Dame}
    \state{IN}
    \country{USA}
}

\author{Max Kreminski}
\affiliation{%
  \institution{Midjourney}
  \city{San Francisco}
  \country{USA}}
\email{mkreminski@midjourney.org}

\renewcommand{\shortauthors}{Chung et al.}

\begin{abstract}

People have different creative writing preferences, and large language models (LLMs) for these tasks can benefit from adapting to each user's preferences. However, these models are often trained over a dataset that considers varying personal tastes as a monolith. To facilitate developing personalized creative writing LLMs, we introduce LiteraryTaste, a dataset of reading preferences from 60 people, where each person: 1) self-reported their reading habits and tastes (stated preference), and 2) annotated their preferences over 100 pairs of short creative writing texts (revealed preference). With our dataset, we found that: 1) people diverge on creative writing preferences, 2) finetuning a transformer encoder could achieve 75.8\% and 67.7\% accuracy when modeling personal and collective revealed preferences, and 3) stated preferences had limited utility in modeling revealed preferences. With an LLM-driven interpretability pipeline, we analyzed how people's preferences vary. We hope our work serves as a cornerstone for personalizing creative writing technologies.\footnote{The dataset can be found in \url{https://github.com/mj-storytelling/LiteraryTaste}}
\end{abstract}



\maketitle

\section{Introduction}
\label{sec:intro}

\begin{figure}
    \centering
    \includegraphics[width=0.478\textwidth]{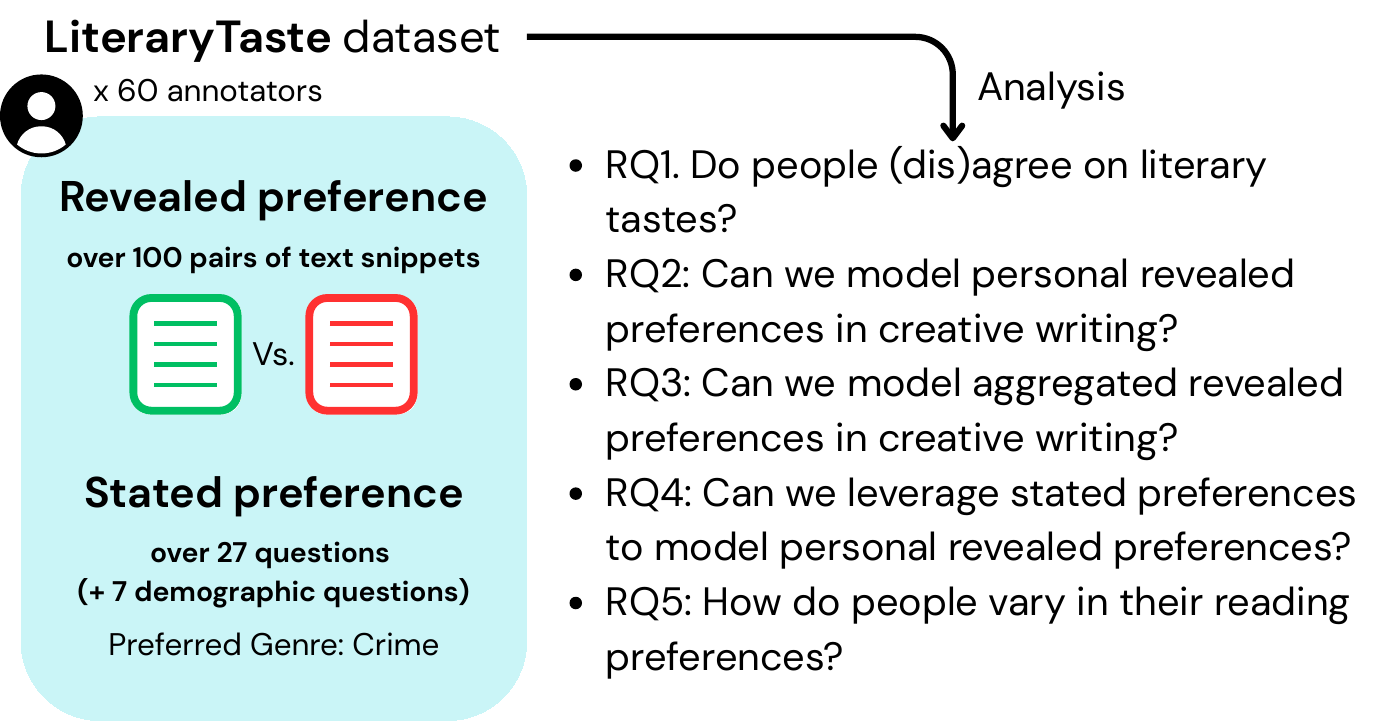}
    \caption{We present \dataset{}, a dataset for creative writing personalization. 60 annotators created the dataset, where each provided 100 binary preference annotations (revealed preference) and 34 survey responses, including those about reading habits and tastes (stated preference). Using the dataset, we addressed a series of research questions related to creative writing personalization.}
    \Description{The left side of the figure is showing LiteraryTaste dataset, which is collected from 60 annotators. Each annotator provided revealed preference, which is binary preference annotations over 100 pairs of text snippets, and stated preferences, where they answered 27 questions on their reading preference and 7 demographic questions. An example of stated preference is "Preferred Genre: Crime." The right side is showing a set of research questions that are answered by analyzing the dataset. The following are research questions: RQ1. Do people have personal reading tastes? RQ2. Can we model personal revealed preferences in creative writing? RQ3. Can we model aggregated revealed preferences in creative writing? RQ4. Can we leverage stated preferences to model personal revealed preferences? RQ5. How do people vary in their reading preferences?}
    \label{fig:teaser}
\end{figure}

Rapid advancements in the text generation capabilities of LLMs have created opportunities for incorporating them in various creative writing applications.
For example, researchers and practitioners are introducing new writing assistants~\cite{lee2024design, chung2024patchview, yuan2022wordcraft, sudowrite} and exploring new forms of generative creative writing media~\cite{kim2024authors, yi2025whatelse, aidungeon}. 
However, the general consensus within the research community is that these models are still far from generating diverse and high-quality creative writings~\cite {chakrabarty2024art, chakrabarty2025can, chung2025modifying}. 

One commonly reported problem is that LLMs tend to produce homogenous texts~\cite{chakrabarty2024creativity, chung2025modifying, anderson2024homogenization}. A  reported cause is that these models are trained with monolithic reward models that learned from aggregate annotator preferences~\cite{ouyang2022training, padmakumar2024beyond}. This averaging could have collapsed the spectrum of preferences into a narrow bin, leading the model to lose personal nuances~\cite{kirk2024benefits}. Considering individual preferences vary greatly for creative writing~\cite{adorno1997aesthetic, bourdieu1984distinction}, homogenized LLM-generated creative writing might not resonate with personal tastes.

To enable LLMs to tailor their outputs to the literary qualities and tastes of different people, we need a way to accurately model this variance in preference. While sociological and aesthetic theory broadly suggest that literary judgments can vary between individuals to some extent \cite{bourdieu1984distinction,barthes1977death,derrida1976grammatology}, it remains unclear how existing technical approaches model literary preferences varying across different users and how much commonalities exist between them. 
We research this question by creating a dataset of various annotators' preferences on creative writings. 
We introduce the \dataset{} dataset, where we collected 60 annotators' 1) \textit{stated preferences}, where they answered a series of survey questions about their reading tastes and behaviors, and 2) \textit{revealed preferences}, where, over 100 pairs of short literary texts, they annotated their preferred one. 

With the dataset, we ran a series of analyses to extend our understanding of creative writing personalization. First, we found that people only minimally agree with each other in their stated and revealed preference responses, which indicates personal tastes exist in both self-aware and implicit ways (RQ1 in Section~\ref{sec:rq1}). Then, we evaluate how different technical approaches model revealed preferences. We found that, among examined approaches (including various LLM-prompted baselines on frontier models), finetuning a transformer encoder (i.e., ModernBERT~\cite{warner2025smarter}) to create a personalized reward model could achieve the best accuracy of 75.8\% when provided 90 revealed preference samples as training data (RQ2 in Section~\ref{sec:rq2}). Moreover, this approach could achieve around 70\% accuracy even with 15 samples, indicating that sample-efficient personal preference modeling is feasible. We also investigated how modeling approaches would perform on aggregated revealed preferences, where we aggregated annotator responses with majority voting. We found that the finetuned transformer encoder still performed the best with 67.7\% accuracy, but it was on-par with prompting an LLM without any input data we collected (RQ3 in Section~\ref{sec:rq3}). 

We then investigated whether it is feasible to incorporate stated preferences in model training, where a single model predicts varying annotators' revealed preferences given their stated preferences. Our results showed that, while stated preferences had some information relevant to inferring revealed preferences, modeling with both types of data did not necessarily lead to better modeling accuracy compared to modeling only with one person's revealed preference (RQ4 in Section~\ref{sec:rq4}). 

Lastly, to analyze characteristics of annotator preferences, we ran an LLM-powered large-scale text analysis pipeline building upon previous work~\cite{lam2024concept}. Through the analysis, we identified 13 key dimensions where annotator preferences vary, and found that annotator clusters exhibit differences along these dimensions. (RQ5 in Section~\ref{sec:rq5}). 
In the discussion, along with interpretations of results, limitations, and future work, we suggest a guideline for designing preference elicitation interactions for personalizable creative writing technologies. 
We release \dataset{} for use by the research community and intend for our analyses to serve as a stepping stone for future work on personalizable creative writing technologies.
\section{Related Work}
We review three categories of previous work: 1) theories and experiments behind how people prefer one creative writing piece over another, 2) human- and machine-based evaluations on creative writings, and 3) personalizing LLM technologies. 

\subsection{Perception of Arts and Creative Writing}
The question of aesthetic preference has been shaped by intellectual movements across centuries. In the eighteenth century, Baumgarten framed aesthetics as a mode of sensorial and imaginative cognition \cite{baumgarten2022reflections}---a foundation that Kant expanded, arguing that aesthetic judgments aspire to universality through a shared human faculty of judgment~\cite{kant2007critique}. In the nineteenth century, Romantic thinkers emphasized the interplay of formal qualities, emotion, and perception \cite{schiller2016}, while late-century Neo-Kantian philosophers and early psychophysicists such as Fechner sought to quantify aesthetic responses \cite{fechner1876}. In the twentieth century, psychological and experiential accounts highlighted the role of subjective experience \cite{freud2003uncanny,dewey1958art}, while mid-century structuralist approaches examined how aesthetic value and taste are structured by underlying cultural systems and conventions \cite{barthes1972mythologies}. Later, poststructuralist \cite{barthes1977death,derrida1976grammatology} and sociological approaches \cite{bourdieu1984distinction} demonstrated how taste is shaped by cultural and social contexts. These traditions suggest that aesthetic preference arises from the interplay between an artwork's properties and the cognitive, cultural, and social frameworks of its audience \cite{liu2006rendering}.

Psychology researchers have conducted empirical experiments that support the above theories. In addition to domains like visual arts~\cite{hagtvedt2008perception} and music~\cite{fredrickson1995comparison}, creative writing has been one of the focal domains of the experiments. Aligned with the above arguments, experiments have shown that both the qualities of the creative writings~\cite{hoshi2018eye, hartung2021aesthetic} and the audience traits and backgrounds~\cite{mak2020influence} impact the appreciation of creative writings. Researchers also found that textual stimuli could impact readers’ psychophysiological responses, providing evidence to support the argument that art appreciation is a psychological experience~\cite{hoshi2018eye, hartung2021aesthetic, wassiliwizky2017theemotional}. While these findings ground our research, no research has yet collected and analyzed personal textual preferences on a large scale of text data, which would help train AI models aware of individual tastes in creative writings. In this research, we collect and analyze large-scale creative writing preference annotations on pairs of text snippets with differing literary styles and content. 

\subsection{Evaluating Creative Writing}
Natural language processing (NLP) and Human-Computer Interaction (HCI) researchers have evaluated creative writing for the purpose of assessing generated texts. Human evaluation has been frequently adopted, either with crowdsourcing~\cite{chhun2024language, xie2023next} or hiring domain experts~\cite{chakrabarty2024art}. Researchers adopted various evaluation structures, such as asking annotators to provide scores for specific criteria (e.g., fluency, interestingness)~\cite{chhun2024language, xie2023next, chung2025toyteller} or to compare a pair of texts regarding those criteria~\cite{chung2022talebrush}. Tian et al.~\cite{tian2024large-language} even annotated story turning points and story arcs to compare LLM-generated texts to human-created ones, and Chakrabarty et al.~\cite{chakrabarty2025can} evaluated LLM-generated texts by asking experts to create their improved versions. More recently, researchers started using LLMs to evaluate the quality of creative writing. While some researchers used general instruction-tuned models~\cite{wu2025longwriterzero, bai2025longwriter}, it has been shown that these models have limitations in their evaluation capabilities~\cite{chakrabarty2024art, chakrabarty2025can}. LLMs finetuned specifically for creative writing evaluation could be a solution for such limitations, and researchers have introduced various approaches to train those models~\cite{wu2025writingbench, chakrabarty2025ai, chen2022storyer, fein2025litbenchbenchmarkdatasetreliable}. Many of these, however, assume that there is one evaluative perspective. Marco et al.~\cite{marco2025reader} recently showed that people’s evaluations of creative writing qualities can vary based on their profiles, such as expertise. While automatic evaluation approaches like WritingBench~\cite{wu2025writingbench} could ideally support personalization with query-specific evaluation, to our knowledge, those approaches only leverage synthetically generated data instead of using real people’s data that genuinely reflects their tastes. We aim to collect various people’s personal preference judgments to facilitate personalized evaluation of creative writing. 

\subsection{Personalizing LLMs}

One thread of work investigated personalizing LLMs by adapting reward models to each user. One of the earliest approaches is Personalized Soups~\cite{jang2023personalized}, which combines multiple predefined rewards with weights during post-training to indicate a user's personal preference. More technically complex approaches followed, such as Poddar et al.~\cite{poddar2024personalizing}, which derived a user's personal latent from their preference data and appended it as a part of the input to the reward model training. In a social group modeling context, PrefPalette~\cite{li2025prefpalette} explored incorporating latent attributes that a specific group might prefer as a part of reward modeling features. While not specifically in the reward modeling context, Orlikowsk et al.~\cite{orlikowski2025beyond} investigated whether we can finetune LLMs to simulate annotations from a specific user with their demographic information. They found demographics helpful, but it was more by memorizing a specific annotator rather than learning demographic patterns. Another thread of work did not encode personal information into the reward model, but as LLM contexts. These approaches often extend direct preference optimization (DPO)~\cite{rafailov2023direct}, which post-trains directly on pairs of winning and losing instances, not on reward models. Li et al.~\cite{li2024personalized} trained LLMs that incorporate user embeddings as LLM contexts, where embeddings are derived from user models co-trained with LLMs on diverse users' preference pairs. Fspo~\cite{singh2025fspo}, similarly, post-trained LLMs with the user's few-shot preference data included within a context. Shaikh et al.~\cite{shaikh2025aligning} investigated an alternative personalization approach, where they leveraged a few user demonstrations (e.g., edits to LLM-generated samples) as a signal to tune LLMs to the user preferences. Recent work started to consider the user's dynamic contexts for personalization, proposing benchmarks~\cite{kim2025cupid} or suggesting prompting-based approaches~\cite{shaikh2025gums}. Previous efforts often rely on synthetic datasets, as the community lacks personalization datasets from real users. Moreover, creative writing has been neglected as a domain for personalization. Hence, we introduce \dataset{} to facilitate research for personalizing LLMs in creative writing.

\section{Collecting \dataset{}}
To facilitate research on personalization in creative writing, we collected data on 1) revealed preferences, or people’s preferences after they read specific creative writings, and 2) stated preferences, or survey questions on their self-reported reading habits and preferences. Revealed preferences show how individuals actually prefer one text over another and can be used to train personalized models. We also collected stated preferences to see if we could infer revealed preferences on specific texts with stated preferences. We explain how we designed our data collection.

\subsection{Revealed Preference Task Design}

While there can be many different aspects of creative writing preferences, as the first step, we focus on understanding those that could be perceived from short text snippets. Accordingly, we focus on two specific aspects: 1) writing style, or how the text is written, which can be defined as each author’s unique writing habits, such as word choice, sentence structure, and paragraph structure~\cite{sebranek1996writers}, and 2) content, or what the text is about.\footnote{Hence, we do not consider aspects that exist in longer texts, such as narrative arcs.} 

We collect revealed preferences via a binary choice task, where annotators chose a preferred text from a pair of texts. We selected this method over fine-grained Likert-scale ratings, which could cognitively overload annotators, or asking them to evaluate more specific aspects (e.g., writing styles), which may be highly varied~\cite{bourdieu1984distinction, adorno1997aesthetic, mak2020influence} or difficult for non-experts to articulate. With binary preference tasks, annotators can make simpler decisions while implicitly considering their own criteria.\footnote{Due to its strengths, the binary preference task is widely used in preference annotation tasks~\cite{christiano2017deep, zheng2023judging}.}

\subsubsection{Annotated Texts}

As we wanted to see whether creative writing preference varies among readers, we covered as diverse a range of creative writing as possible with five datasets:
\begin{itemize}
    \item \textbf{Gutenberg} dataset is a collection of copyright-free books from Project Gutenberg.\footnote{https://www.gutenberg.org/} The majority of these have been published before the 2000s. We used a Huggingface dataset repository that filtered fiction texts (392080 instances).\footnote{https://huggingface.co/datasets/sanps/GutenbergFiction}
    \item \textbf{Sterman et al.}~\cite{sterman2020interacting} have collected freely available book previews of modern fiction from Amazon Kindle. The dataset reflects modern literary writing styles (1729 instances). 
    \item \textbf{r/WritingPrompts}~\cite{fan2018hierarchical} is an online forum where users share creative writings according to writing prompts provided by others. This dataset reflects modern writing style, not necessarily by professional writers (598651 instances). 
    \item \textbf{Poetry dataset}\footnote{https://huggingface.co/datasets/merve/poetry} is a collection of Renaissance or modern poetry on topics of a) love, b) nature, and c) mythology and folklore (308 instances). The data originates from the Poetry Foundation website.\footnote{https://www.poetryfoundation.org/} 
    \item \textbf{Tell-me-a-story}~\cite{huot2025agents} is a short story collaboratively created by expert creative writers for evaluation purposes, with a workshop that involves initial drafting, receiving feedback, and revising (123 instances).
\end{itemize}

After cleaning, for Poetry, Tell-me-a-story, Sterman et al.’s datasets, we sampled 308, 123, and 1014 instances, respectively. For Gutenberg and r/writingPrompt, we sampled 1063 and 1092 instances, respectively. These resulted in a total of 3600 instances.\footnote{The counts are after cleaning duplicates and non-creative writings, such as table of contents or author comments.} For each text instance, we randomly sampled a 150-word snippet, as our focus was on seeing people’s preferences over short texts.

In addition to instances from the dataset, researchers could be curious to compare with LLM-written texts. Hence, we included pairs where we compare human-written texts with LLM-generated texts, 1) on different contents and 2) with the same topics. Comparisons of 2) would specifically reveal preferences over the writing styles of humans and LLMs. For 1), we sampled 200 text snippets from 3600 instances and replaced them with LLM-generated texts on the same topic. For 2), we sampled 400 text instances and paired them with LLM-generated texts on the same topic. To generate texts, following \citet{chakrabarty2024art}, we first extracted topics or themes of those snippets with an LLM, and then prompted LLMs to write literary texts on extracted topics. We used \texttt{claude-4-sonnet} and \texttt{GPT-4.1}, with each generating half of the required instances. After generating texts, for snippets other than those already paired in 2), we randomly paired them with human-written texts. This resulted in 1400 pairs of both human-authored texts, 200 pairs with human-written and LLM-generated texts on different topics, and 400 pairs with human-written and LLM-generated texts on the same topic (total 2000 pairs). 

\begin{table*}[]
\caption{Stated preference questions. Note that open-ended responses (``Other'') were possible for multiple selection questions.}
\adaptivetablesize
\centering
\begin{tabular}{p{0.08\textwidth}p{0.18\textwidth}p{0.29\textwidth}p{0.35\textwidth}}
\thickhline
Type & \multicolumn{2}{l}{Question} & Answer format \\
\thickhline
\multirow{15}{0.08\textwidth}{Reading frequency \cite{throsby2017australian}} & \multicolumn{2}{p{0.50\textwidth}}{Approximately, how many hours do you spend reading for pleasure per week?} & Number \\
\cline{2-4}
& \multicolumn{2}{p{0.49\textwidth}}{Approximately, how many hours do you spend reading per week (including hours reading for pleasure)?} & Number \\
\cline{2-4}
& \multicolumn{2}{p{0.49\textwidth}}{How many books do you read per month?} & Number \\
\cline{2-4}
& \multirow{11}{0.18\textwidth}{How frequently do you enjoy the following leisure activity?} & Watch videos (e.g., TV, YouTube, or Netflix) & \multirow{11}{0.35\textwidth}{Ordinal (Every day, At least once per week, Less often than once per week, Never)} \\
\cline{3-3}
& & Read textual content (e.g., books, web serial fictions, articles) & \\
\cline{3-3}
& & Exercise (not sport, e.g., gym workouts, running, cycling, yoga) & \\
\cline{3-3}
& & Creative craft activity (e.g., knitting, woodworking, jewelry making) & \\
\cline{3-3}
& & Play video games & \\
\cline{3-3}
& & Creative art activity (e.g., painting, creative writing, music performance/composition) & \\
\cline{3-3}
& & Play sports & \\
\hline
\multirow{16}{0.08\textwidth}{Reading motivation \cite{throsby2017australian}} & \multirow{16}{0.18\textwidth}{Choose specific reasons why you read.} & \multirow{7}{0.29\textwidth}{Enjoyment} & 
{Multiple selection (Be inspired/uplifted, Be part of a conversation about books, Drama of good stories / watch a good plot unfold, Escape reality / become immersed in another world, Pass the time / keep myself occupied, Read good writing, Read to another person, Spend time with my favorite authors/characters/settings, Stimulate my imagination and creativity)}
\\
\cline{3-4}
& & \multirow{7}{0.29\textwidth}{Learning} & 
{Multiple selection (Engage with literature and ideas, Expand my world view, Help me reflect on / deal with life's challenges, Improve my analytical/critical thinking, Improve my writing, Improve reading speed / learn new words, Learn about the world through other people's experiences, Learn about topics that interest me, Learn or improve practical skills)}
\\
\cline{3-4}
& & \multirow{2}{0.29\textwidth}{Health and others} & 
{Multiple selection (For company, For relaxation/stress release, Help sleep, Improve mental health)}
\\
\hline
\multirow{8}{0.08\textwidth}{Reading genre \cite{throsby2017australian}} & \multirow{8}{0.18\textwidth}{Which of the following types of texts do you enjoy reading these days?} & \multirow{3}{0.29\textwidth}{Fictions} & 
{Multiple selection (Classics, Contemporary/general fiction, Crime/mystery/thriller, Historical, Literary, Poetry, Romance, Science fiction/fantasy, Young adult/teen)}
\\
\cline{3-4}
& & \multirow{5}{0.29\textwidth}{Non-fictions} & 
{Multiple selection (Autobiography/biography/memoir, Cookbooks/food and drink, Crafts/hobbies/pets, Gardening/home improvement, Health/diet/wellbeing, History, Humour, Personal Development, Science \& Nature, Social \& Cultural, Travel guides/travel narratives)}
\\
\hline
\multirow{12}{0.08\textwidth}{Preferred textual qualities \cite{vakkari2019theroleof}} & \multirow{12}{0.18\textwidth}{Rate how important below factors are when you are reading texts.} & Arouses feelings & \multirow{12}{0.35\textwidth}{Ordinal (Not important at all, Slightly important, Somewhat important, Very important)} \\
\cline{3-3}
& & Based on real events & \\
\cline{3-3}
& & Challenges the reader & \\
\cline{3-3}
& & Entertaining & \\
\cline{3-3}
& & Gripping content / plot & \\
\cline{3-3}
& & Original style & \\
\cline{3-3}
& & Represent facts and reality & \\
\cline{3-3}
& & Rich characters & \\
\cline{3-3}
& & Fiction's setting precisely presented in detail & \\
\cline{3-3}
& & Skillful and rich language & \\
\cline{3-3}
& & Surprising content / plot & \\
\cline{3-3}
& & Thought provoking & \\
\thickhline
\end{tabular}
\label{tab:stated_preference_questions}
\Description{The first section focuses on reading frequency, containing four high-level questions. First, it asks approximately how many hours you spend reading for pleasure per week, requesting a number as the answer. Second, it inquires about total weekly reading hours, including both pleasure reading and other reading, also requesting a number. Third, it asks how many books you read per month, again requesting a numerical response. The fourth explores leisure activity frequency, asking how often you enjoy various activities, with a sub-question per activity. The response options are ordinal choices: every day, at least once per week, less often than once per week, or never. The activities listed include watching videos such as TV, YouTube, or Netflix; reading textual content like books, web serial fiction, or articles; exercise activities including gym workouts, running, and cycling; creative craft activities such as knitting, woodworking, or jewelry making; playing video games; creative art activities including painting, creative writing, and music performance or composition; and playing sports. The second section examines reading motivation through two questions. The first asks you to choose specific reasons why you read, with multiple selection options. The enjoyment category includes various reasons such as being inspired or uplifted, being part of a conversation about books, experiencing drama of good stories or watching a good plot unfold, escaping reality or becoming immersed in another world, passing the time or keeping yourself occupied, reading good writing, reading to another person, spending time with favorite authors, characters, or settings, and stimulating imagination and creativity. The learning category offers options including engaging with literature and ideas, expanding your world view, helping to reflect on or deal with life's challenges, improving analytical or critical thinking, improving writing skills, improving reading speed or learning new words, learning about the world through other people's experiences, learning about topics that interest you, and learning or improving practical skills. The health and others category includes options for company, relaxation or stress release, helping sleep, and improving mental health. The next question asks which types of texts you enjoy reading these days, with multiple selection options for both fiction and non-fiction. Fiction genres include classics; contemporary or general fiction; crime, mystery, or thriller; historical; literary; poetry; romance; science fiction or fantasy; and young adult or teen. Non-fiction categories include autobiography, biography, or memoir; cookbooks, food and drink; crafts or hobbies or pets; gardening or home improvement; health, diet, wellbeing; history; humor; personal development; science and nature; social and cultural; and travel guides or travel narratives. The fourth and final section addresses preferred textual qualities, asking you to rate the importance of various factors when reading texts. The rating scale is ordinal: not important at all, slightly important, somewhat important, or very important. The qualities to rate include whether the text arouses feelings, is based on real events, challenges the reader, is entertaining, has gripping content or plot, is in an original style, represents facts and reality, has rich characters, has a fiction setting precisely presented in detail, uses skillful and rich language, has surprising content or plot, and is thought provoking.}
\end{table*}

\subsection{Stated Preference Task Design}
We designed our survey to identify people’s reading habits and preferences with a self-report questionnaire (Table~\ref{tab:stated_preference_questions}). Grounding our survey questions with previous research~\cite{throsby2017australian, vakkari2019theroleof}, we focused on identifying reading frequency, reading motivation, reading genre, and reading preference.\footnote{While we started from the questions from the previous work, we revised the questions as necessary with pilot studies.} 

For reading frequency, we asked about the number of hours spent reading, both in total hours and in reading for pleasure. We also asked about the number of books they read per month. Moreover, we included questions about the frequency of their leisure activities, including and beyond reading (e.g., playing sports, creating arts). These frequency questions had four options: every day, at least once per week, less often than once per week, and never.

Reading motivation questions asked for reasons why they read. Participants could select multiple options while being allowed open-ended responses. We put them into three high-level categories: enjoyment (e.g., be inspired or uplifted), learning (e.g., expand my world view), and health (e.g., improve mental health). 

We asked for preferred reading genres with two questions, one on fiction genres (e.g., romance) and the other on non-fiction ones (e.g., history). For each question, people could select multiple options or provide open-ended responses.

For reading preferences, we asked which textual characteristics they care about. We provided 12 aspects, and we asked people to rate each of them with a four-point scale: not important at all, slightly important, somewhat important, and very important.

\begin{figure*}
    \centering
    \includegraphics[width=\textwidth]{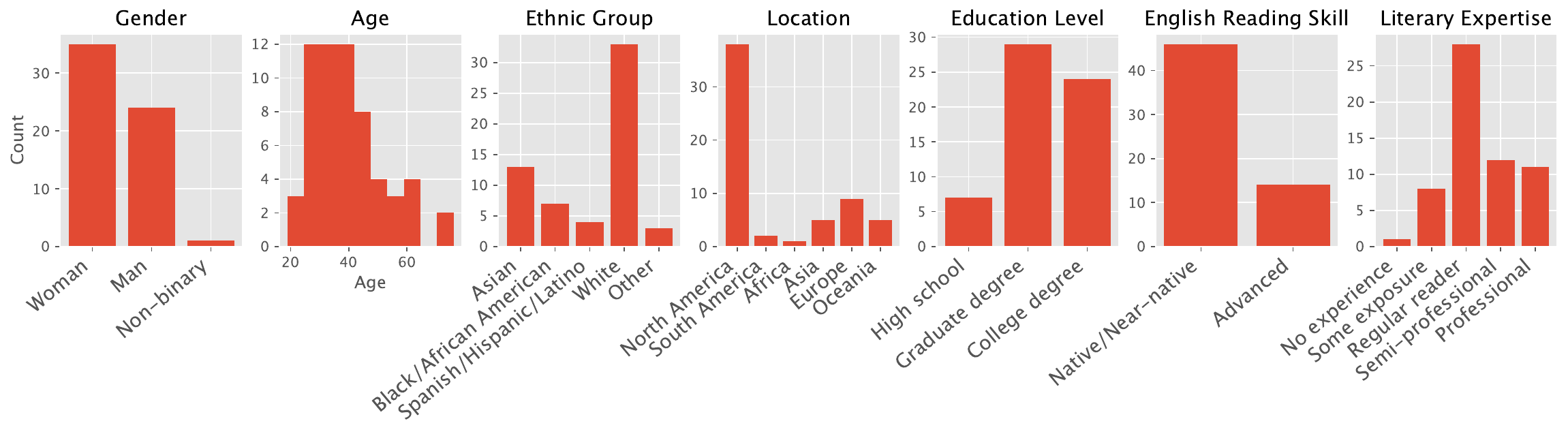}
    \caption{Demographics of data collection participants.}
    \label{fig:demographic}
    \Description{This figure shows demographic information about study participants across seven bar charts. The Gender chart reveals approximately 35 women, 24 men, and 2 non-binary participants. The Age distribution chart shows participants primarily between 20 and 60 years old, with the highest concentration in the 30-40 age range, peaking at about 12 participants around age 30, and smaller numbers at the extremes with 3 participants around age 20 and 2 around age 70. The Ethnic Group chart indicates White participants form the largest group at about 33 people, followed by Asian participants at approximately 13, with smaller representations of Black/African American at 7, Hispanic/Latino at 4, and Other at 3. The Location chart shows North America as the dominant region with more than 35 participants, followed by Europe with around 9, Asia with around 5, Oceania with around 5, South America with around 3, and Africa with around 2. For Education Level, the Graduate degree category has the highest representation at approximately 28 participants, followed by College degree at 24, with High school having about 7 participants. The English Reading Skill chart shows the vast majority, about 47 participants, identify as Native/Near-native speakers, with Advanced speakers at 13 participants. Finally, the Literary Expertise chart demonstrates most participants as regular readers at about 29 people, followed by about 12 semi-professionals, about 11 professionals, about 8 people with some exposure, and 1 with no experience.}
\end{figure*}

\subsection{Data Collection Procedure}
We hired participants from Upwork, those who are 1) living in English-speaking countries and 2) have an above 90\% task success rate. We tried to cover various demographic traits and levels of reading experiences (Figure~\ref{fig:demographic}). While hosting the task with Potato~\cite{pei2022potato} (the screenshot in Appendix~\ref{app:collection_interface}), each participant completed the survey questions for demographics and stated preferences first and then annotated their revealed preferences over 100 pairs of texts. 
We placed stated preference questions first, as exposure to actual creative writings could bias their preconception of their reading taste.
As all tasks for revealed and stated preferences could take two to five hours, we paid participants \$100 (\$20 to \$50 hourly payments). For the initial 15 participants, we collected data without attention checks, but realized that two of them completed the task very quickly (<10 seconds per text pair), seemingly paying no attention to the tasks. Hence, while filtering out those two, for the later participants, we added two attention checks within the task. With the attention checks, we collected 48 participants’ data, and filtered out one who did not pass the attention checks. Note that three participants annotated each set of 100 pairs of texts. In total, we collected data from 60 people over 2000 text pairs.
\section{Analysis on \dataset{}}
We answer a series of research questions by analyzing the dataset.

\subsection{RQ1. Do people (dis)agree on literary tastes?}
\label{sec:rq1}

\subsubsection{Motivation}
The first research question focused on confirming that people have different tastes regarding creative writing. 

\subsubsection{Analysis Method}
We evaluated the agreement between annotators.
For revealed preferences, we computed Fleiss’ Kappa~\cite{fleiss1971measuring} for those who annotated the same set of text pairs. 
For the agreement of stated preferences, we calculated Krippendorf's alpha as the survey has mixed data types~\cite{marzi2024k}. Specifically, we calculated the alpha scores per the combination of Type and Answer Format (see Table~\ref{tab:stated_preference_questions}). We mapped the Answer Format of Number (real values) to ratios, where the maximum of the response serves as 100\%. For Multiple selection questions, we considered each option as a binary nominal category. As a result, we got alphas for \texttt{Reading frequency\allowbreak-Number}, \texttt{Reading frequency\allowbreak-Multiple selection}, \texttt{Reading motivation\allowbreak-Nominal}, \texttt{Reading genre\allowbreak-Multiple selection}, and \texttt{Preferred textual qualities\allowbreak-Ordinal}.

\subsubsection{Results}
For revealed preferences, the average Fleiss’ Kappa was \textbf{0.1405} with a standard deviation of 0.1005. It indicates slight or poor agreement between annotators~\cite{fleiss2013statistical}, which we interpret as that, \textbf{while people agreed on preferring certain textual qualities, they still had some disagreements with each other}. The minimum Kappa was -0.0179, and the maximum was 0.3805. 

\begin{table}[]
\caption{Krippendorf's alpha for stated preference questions}
\begin{tabular}{l|l}
\hline
\texttt{Reading frequency-Number}            & 0.3133  \\
\texttt{Reading frequency-Ordinal}           & 0.4694 \\
\texttt{Reading motivation-Multiple selection}          & 0.1490 \\
\texttt{Reading genre-Multiple selection}               & 0.0620 \\
\texttt{Preferred textual qualities-Ordinal} & 0.1789 \\ \hline
\end{tabular}
\label{tab:stated_pref_agreement}
\Description{This table presents Krippendorff's alpha values for five stated preference questions measuring inter-rater reliability. The Reading frequency question coded as a number shows an alpha of 0.3133, while the same question coded as ordinal data yields a higher reliability of 0.4694. The Reading motivation question using multiple selection format has an alpha of 0.1490, and the Reading genre question, also using multiple selection, shows the lowest reliability at 0.0620. The Preferred textual qualities question coded as ordinal data has an alpha of 0.1789.}
\end{table}

For the stated preference questions, we present the alpha results in Table~\ref{tab:stated_pref_agreement}. As shown, people had higher agreement on reading frequency than on other question types. Other than reading frequency, all alpha values were below 0.2. As alpha values higher than 0.6 are considered as acceptable agreement~\cite{marzi2024k}, we conclude that \textbf{people only slightly agree on their stated preferences}.

\subsection{RQ2: Can we model personal revealed preferences in creative writing?}
\label{sec:rq2}

\subsubsection{Motivation}

As the second research question, we were curious about whether we can model each person's revealed preferences with existing technical approaches. If we can reliably model each person's revealed preferences, it would give many opportunities for personalizing LLMs in the creative writing domain. For example, we can use such models as reward models to finetune LLMs to different tastes. As the first step, we investigate whether we can model each person's revealed preference only by using that person's revealed preference data.

\subsubsection{Analysis Method}
For the analysis, we focused on modeling the task that the annotators did: deciding the preferred text snippet from a pair of them. 
This could be effective for creating datasets for some post-training approaches, such as Direct Preference Optimization (DPO)~\cite{rafailov2023direct}. 
For the model developer who wants to have numerical rewards, for example, to use Group Relative Policy Optimization (GRPO)~\cite{shao2024deepseekmath}, they can still indirectly calculate it, such as computing Elo scores~\cite{elo1978rating} from binary preferences. 

While we examined a variety of technical approaches, we evaluated their performance by running 10-fold validations on the revealed preference annotations. That is, while running training 10 times, for each fold, 10 annotations served as a test set and 90 as a training set. We calculated accuracies over 10 training runs over 60 annotators (in total, 600 accuracy results). Note that, as some of our modeling approaches do not allow ``unsure'' labels, when computing the accuracy, we did not consider instances with ``unsure'' labels.

\begin{figure*}
    \centering
    \includegraphics[width=\textwidth]{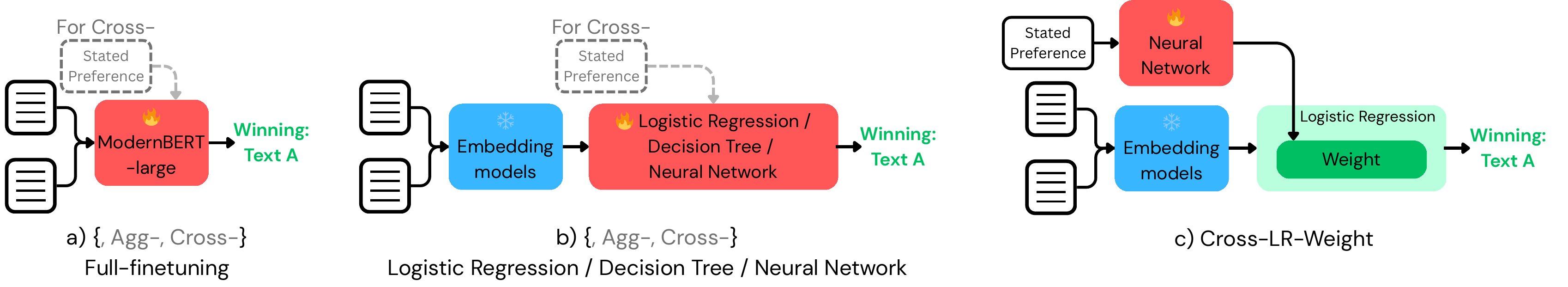}
    \caption{Training approaches in RQ2, 3, and 4. Red, blue, and green indicate tuned weights, frozen models, and model output, respectively. a) For \texttt{Full-Finetuning}-based approaches, we finetuned all weights of the transformer encoder. b) For \texttt{Logistic Regression}, \texttt{Decision Tree}, and \texttt{Neural Network}-based approaches, we first embedded texts with frozen embedding models and then trained corresponding models with embeddings as training inputs. Approaches in a) and b) could be trained for aggregated preferences (\texttt{Agg-}, in RQ3) and \texttt{Cross}-annotator models (i.e., taking stated preference input to infer the preference from the perspective of annotators who would have such stated preference, in RQ4). c) \texttt{Cross-LR-Weight} (in RQ4) trains a neural network model that infers the weight of a logistic regression model given stated preference input. Note that, as embedding models, we used \texttt{jinaai/jina-embeddings-v4}~\cite{günther2025jinaembeddingsv4} and ModerBERT-large finetuned on the style similarity dataset~\cite{sterman2020interacting}.}
    \label{fig:modeling_approaches}
    \Description{This figure illustrates the conceptual architecture of different training approaches used in Research Questions 2, 3, and 4, with color coding where red indicates tuned weights, blue represents frozen models, and green shows model outputs. Panel (a) demonstrates the full-finetuning approach for personal, aggregated, and cross-annotator preferences. Two text documents feed into ModernBERT-large shown in red, indicating all model weights are trainable. The model directly outputs a winning text prediction, either Text A or B. For cross-annotator experiments, the dashed box indicates the model receives stated preference information as additional input to condition its predictions. Panel (b) shows the approach for Logistic Regression, Decision Tree, and Neural Network models, also applicable to personal, aggregated, and cross-annotator settings. Text documents first pass through embedding models shown in blue, indicating these are frozen pre-trained models that convert text to numerical representations. These embeddings then feed into trainable classification models shown in red, including Logistic Regression, Decision Tree, or Neural Network architectures. The models output winning text predictions. For cross-annotator variants, stated preferences again provide additional conditioning information as indicated by the dashed box. Panel (c) depicts the Cross-LR-Weight approach introduced in Research Question 4. This architecture combines stated preferences with text embeddings in a more sophisticated manner. Text documents pass through frozen embedding models in blue, while stated preferences feed into a separate Neural Network shown in red. This neural network learns to predict appropriate weights for a Logistic Regression model. The embedding models' outputs and the weight predictions converge at a Logistic Regression layer shown in green with "Weight" label, which then produces the final text preference prediction.}
\end{figure*}

We evaluated a variety of modeling approaches as follows:

\paragraph{Finetuned ModernBERT-large (\texttt{Full-Finetuning})} 
We finetuned all weights of an encoder transformer, ModernBERT-large~\cite{warner2025smarter} (Figure~\ref{fig:modeling_approaches}a), as a reward model ($r_{\theta}(x)$ where $x$ is a text snippet) with a binary ranking loss~\cite{ouyang2022training}:
\begin{equation}
    \mathcal{L}=-\text{log}(\sigma(r_{\theta}(x_c) - r_{\theta}(x_r)))
\end{equation} 
$x_c$ and $x_r$ are chosen and rejected snippets, respectively. While this model outputs numerical scores, we used those scores to decide which text was preferred from a pair (i.e., considering the text with the higher score as preferred). As this approach only uses pairs with clear winning and losing instances, we did not use pairs where the annotators marked their preference as ``unsure.'' 

We trained \texttt{Full-Finetuning} using Huggingface's Trl library.\footnote{https://huggingface.co/docs/trl/v0.19.1} We used the initial learning rate of 5.0e-5 and batch size of 8. We ran the training for 10 epochs, evaluating after every epoch. We picked the results with the best test accuracy.
    
\paragraph{Logistic regression over embedded texts (\texttt{Logistic Regression\allowbreak-\{All, Sem, Sty\}})} 
We trained logistic regression models over frozen text embedding spaces (Figure~\ref{fig:modeling_approaches}b). These models have a strength that their coefficient can serve as interpretable vectors about the annotator's preference~\cite{kim2018interpretability}. As we consider a pair of texts as input, we embed the texts and then get the directional vector by subtracting one of them from the other~\cite{springall1973paired}. Logistic regression model classified the directional vector, whether the annotator would prefer the subtracting text or the subtracted text. Note that we could consider the ``unsure'' label as the third class. We used two embedding models to embed texts: semantic (\texttt{Sem}) and style (\texttt{Sty}) models. We also considered a condition that uses the concatenation of vectors from two embedding models (\texttt{All}). For semantic embedding, we used \texttt{jinaai/jina-embeddings-v4}~\cite{günther2025jinaembeddingsv4}. For style embedding, we finetuned all weights of the ModernBERT-large with the dataset from Sterman et al.~\cite{sterman2020interacting} (\texttt{style-ModernBERT-large}, See Appendix~\ref{app:styleembedding} for training details and model performance). 
We trained logistic regression models with \texttt{scikit-learn}\footnote{https://scikit-learn.org/}, with max iteration of 1000.
    
\paragraph{Decision tree over embedded texts (\texttt{Decision Tree-\{All, Sem, Sty\}})} 
We similarly trained decision tree models on frozen embedding spaces (of \texttt{jinaai/jina-embeddings-v4} and \texttt{style-Modern\allowbreak BERT-large}) as we trained logistic regression models (Figure~\ref{fig:modeling_approaches}b). We embedded texts from a pair into vectors, subtracted one of them from the other, and then classified the directional vector based on the annotator's preference. These models also have the benefit that we can interpret the model's behavior by visualizing trees. We used \texttt{scikit-learn} to train models, with default parameters.

\paragraph{Neural Network over embedded texts (\texttt{Neural Network\allowbreak-\{All, Sem, Sty\}}} 
Similar to logistic regression and decision tree models, we trained 2-layer neural network models on frozen embedding spaces (Figure~\ref{fig:modeling_approaches}b). While not interpretable, these models have a higher modeling capability than logistic regression and decision trees (while having fewer tunable parameters than \texttt{Full-Finetuning}). We used \texttt{scikit-learn}'s \texttt{MLPClassifier} to train models, with two hidden layers, each having a size of 4096. We used \texttt{tanh} for the activation and the initial learning rate of 3e-4. We trained models for 40 epochs, while evaluating per four epochs, and used the results with the best test accuracy.

\paragraph{Few-shot LLM prompting (\texttt{o4-mini\allowbreak-\{Rand, Sim\}} and \texttt{Sonnet\allowbreak-4\allowbreak-Rand\allowbreak\{, -RSOff\}})}
We examined whether LLMs can discern an annotator's preferred texts with a few shots of examples. We considered two models capable of reasoning~\cite{xu2025largereasoningmodelssurvey}, OpenAI's \texttt{o4-mini}\footnote{o4-mini-2025-04-16} and Anthropic's \texttt{Sonnet-4}.\footnote{claude-sonnet-4-20250514} 
When prompting, we sampled five examples from the training set. It is because putting all 90 training samples in the prompt is practically inefficient due to high token usage. To get the results efficiently, we retrieved results on all 10 test set items with a single LLM call. For specific prompts we used, please refer to Appendix~\ref{app:usedprompts}.

For \texttt{o4-mini}, we tested two approaches to sample few-shot examples: 1) randomly sampling examples (\texttt{Rand}) and 2) sampling examples similar to the queried inputs (\texttt{Sim}). Specifically, when measuring the similarity, for each text pair, we first obtained the directional vector by embedding the texts with the semantic and style models and then subtracting them. Then, we measured the similarity between pairs by calculating the cosine similarity between the obtained directional vectors. Note that, as directions can flip based on which text is used as a subtracting embedding, we used the absolute value of the cosine similarity as the similarity metric. As we had 10 queried inputs per LLM call, we first sampled 10 training instances that are most similar to each of the 10 query inputs, then used five examples with the highest similarity score. 

For \texttt{Sonnet-4}, as we can turn on and off the reasoning capability, we also examined the performance without reasoning (\texttt{RSOff}). Only for \texttt{RSOff}, we used the temperature of 0.

\paragraph{LLM-based profile synthesis (\texttt{o4-mini\allowbreak-Synth})} 
One limitation of few-shot prompting is that we cannot comprehensively consider training samples unless we put them all into the prompt. To overcome this limitation, we examined SynthesizeMe!~\cite{ryan2025synthesizeme}, which synthesizes binary preference annotations into a maximally informative user profile in natural language with bootstrapped LLM reasoning. We used SynthesizeMe! to derive the user profile from 90 training samples and then included it in a prompt to get preference predictions. For the synthesis of the user profile, we used OpenAI's \texttt{gpt-4o-mini}.\footnote{openai/gpt-4o-mini-2024-07-18} Similar to other LLM prompting conditions, we sampled all 10 test set results with a single LLM call. Refer to Appendix~\ref{app:usedprompts} for the prompts we used. For this condition, we only ran the evaluation over five folds, as it took a lot of time and resources to run SynthesizeMe! on 90 training samples.

\paragraph{Varying training set size.}
We were curious how the training set size would impact the model performance. Hence, for \texttt{Full\allowbreak-finetuning}, \texttt{Logistic Regression\allowbreak-All}, \texttt{Decision Tree\allowbreak-All}, \texttt{Neural Network\allowbreak-All}, and \texttt{o4-mini-Synth} we varied the training set size from 15 to 30, 60, and 90, and examined how the test accuracy changes.

\subsubsection{Results}

\begin{figure}
    \centering
    \includegraphics[width=0.478\textwidth]{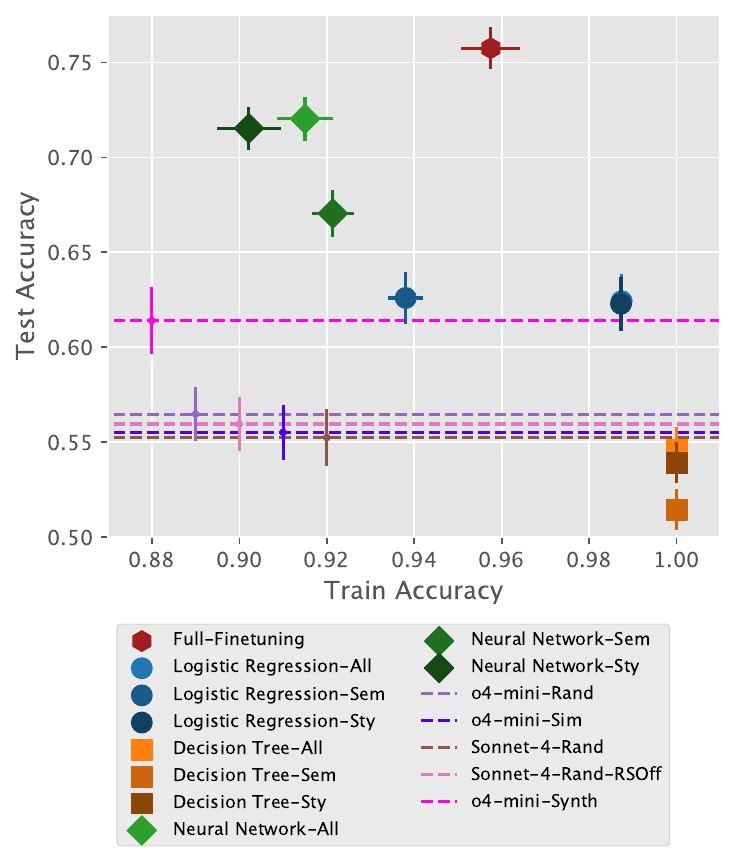}
    \caption{RQ2 results on personal preference modeling. \texttt{All} indicates training models on the concatenation of the semantic and style embeddings of texts, while \texttt{Sem} and \texttt{Sty} indicate only using semantic or style embeddings, respectively. \texttt{Rand} and \texttt{Sim} indicate sampling few shots either randomly or based on sample similarities, respectively. \texttt{RSOff} means turning off reasoning capability, while \texttt{Synth} uses SynthesizeMe!~\cite{ryan2025synthesizeme} to infer user profiles. Note that \texttt{o4-mini} and \texttt{Sonnet-4} approaches do not have training accuracy as they are prompting-based. Error bars and ranges in this paper indicate 95\% confidence intervals.}
    \label{fig:rq2}
    \Description{This figure displays a scatter plot comparing test accuracy against train accuracy for different machine learning models and baselines used in a classification task. The plot shows results for various model types including Full-Finetuning, Logistic Regression, Decision Trees, and Neural Networks. Logistic Regression, Decision Trees, and Neural Network have different feature sets denoted as All, Semantic (Sem), and Stylistic (Sty). The horizontal axis represents train accuracy ranging from 0.88 to 1.00, while the vertical axis shows test accuracy from 0.50 to 0.75. Several horizontal reference lines cross the plot, including a pink dashed line at approximately 0.615 labeled as o4-mini-Synth, a purple dashed line at about 0.56 for o4-mini-Rand, and pink and blue dashed lines around 0.555 for both Sonnet-4-Rand-RSoff and o4-mini-Sim, with another purple line at 0.55 for Sonnet-4-Rand. Full-Finetuning, shown as a red point, achieves the highest test accuracy at approximately 0.76 with a train accuracy near 0.96. The Neural Network models, displayed in green, show varied performance with the Style-features version at 0.71 test accuracy and 0.90 train accuracy, the All version at 0.72 test and 0.91 train, and the Semantic version at 0.67 test and 0.92 train accuracy. Logistic Regression models, represented in blue, cluster in the middle range with test accuracies between 0.62 and 0.63. All and Style variants are achieving similar train accuracies around 0.99. The semantic variant is achieving a train accuracy around 0.94. The Decision Tree models, shown in orange and brown, appear at the lower end of performance with test accuracies between 0.51 and 0.54, despite achieving perfect or near-perfect train accuracy of 1.00, suggesting significant overfitting.}
\end{figure}

Figure~\ref{fig:rq2} shows the analysis results. \textbf{\texttt{Full\allowbreak-Finetuning} had the highest test accuracy}, followed by \texttt{Neural Network\allowbreak-All} and \texttt{Neural Network\allowbreak-Sty}. Among \texttt{Neural Network} approaches, \texttt{Neural Network\allowbreak-Sem} showed the lowest test accuracy. \texttt{Logistic\allowbreak Regression} approaches followed, where different embedding approaches had similar performances. 
Despite evidence from past work about using LLMs for preference modeling \cite{li2025eliciting}, we find that even the strongest frontier models are outperformed by simple supervised methods like logistic regression.
For LLM prompting approaches, only \texttt{o4-mini-Synth} had almost on-par, slightly lower performance than \texttt{Logistic Regression} approaches. The other few-shot prompting approaches all had lower test accuracies, around 0.55 to 0.56. \texttt{Decision Tree} had the lowest test accuracy, with \texttt{Decision Tree\allowbreak-All} performing best among them. \texttt{Decision Tree\allowbreak-Sty} and \texttt{Decision Tree\allowbreak-Sem} followed after in the order.

\begin{figure}
    \centering
    \includegraphics[width=0.478\textwidth]{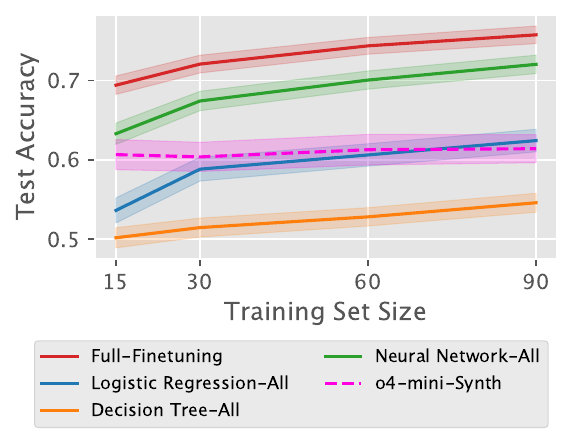}
    \caption{RQ2 results with varying training set sizes.}
    \label{fig:rq2_varying}
    \Description{This figure illustrates how test accuracy changes as training set size increases from 15 to 90 samples for four different models. The graph shows Training Set Size on the horizontal axis and Test Accuracy on the vertical axis ranging from 0.5 to above 0.8, with shaded confidence bands around each line indicating uncertainty. Full-Finetuning, represented by the red line, demonstrates the strongest performance across all training set sizes, starting at approximately 0.69 accuracy with just 15 training samples and gradually increasing to about 0.76 at 90 samples. The model shows steady improvement with more training data, though the rate of improvement decreases as the training set grows larger. Neural Network-All, shown in green, performs second best, beginning at around 0.64 accuracy with 15 samples and rising to approximately 0.72 at 90 samples. This model exhibits a similar learning curve to Full-Finetuning but consistently remains about 0.04 to 0.05 accuracy points below it. Logistic Regression-All, depicted in blue, starts at about 0.54 accuracy with 15 samples and shows the steepest initial improvement, reaching around 0.63 accuracy at 90 samples. The curve shows more pronounced improvement in the 15 to 30 sample range before leveling off. Decision Tree-All, represented by the orange line, shows the poorest performance and least improvement with additional training data. It begins at approximately 0.50 accuracy with 15 samples and only reaches about 0.55 at 90 samples, displaying an almost flat trajectory. o4-mini-Synth has a flat accuracy around 0.61 regardless of training set size.}
\end{figure}

When varying the training set size (Figure~\ref{fig:rq2_varying}), except for \texttt{o4-mini\allowbreak-Synth}, the performance increase was largest between using 15 samples and using 30 samples. The performance increase existed afterward, but the amount of increase was smaller. However, the performance does not saturate with 90 instances, implying that \textbf{if we train models with a size larger than 90 training samples, it would likely produce models with even higher test accuracy}. For \texttt{o4-mini-Synth}, the performance did not change much with varying sizes of training set; possibly because the approach relies on generating natural language user profiles given training inputs. Note that \textbf{\texttt{Full-Finetuning} could achieve around 0.7 test accuracy even with 15 samples}, indicating that finetuning well-pretrained transformers, even with a small sample size, could be effective to achieve high test accuracy.

\subsection{RQ3: Can we model aggregated revealed preferences in creative writing?}
\label{sec:rq3}

\subsubsection{Motivation}
While RQ1 results show that people have varying reading tastes, it also indicates that there are \textit{some} agreements in revealed preferences. Hence, we became curious if we could model ``aggregated'' preferences. If existing technical approaches could model such aggregated preferences, it would mean that there are textual qualities that people universally agreed to prefer, while diverging on other aspects. 

\subsubsection{Analysis Method}
We first aggregated preference annotations among the three people's annotations per text pair. Specifically, we considered that a text is collectively preferred over the other if the number of people who prefer it is higher than that of those who do not prefer it. If annotators collectively preferred neither from a pair, we assigned an unsure label to the pair. The aggregation results in a total of 2000 text pairs.

After the aggregation, we applied the same technical approaches as RQ 2 (Section~\ref{sec:rq2}) to model aggregated preferences. We only did not examine \texttt{Synth} as it is inherently designed to model personal profiles with LLMs~\cite{ryan2025synthesizeme}. Moreover, we added conditions that prompt LLMs in zero-shot, where we asked LLMs to predict which text people would prefer \textit{generally}, without providing examples (\texttt{Zero}). We ran 10-fold validations over the aggregated dataset. Note that we differentiate conditions in this analysis from conditions of the previous study by prepending \texttt{Agg} in their names.

\subsubsection{Results}

\begin{figure}
    \centering
    \includegraphics[width=0.478\textwidth]{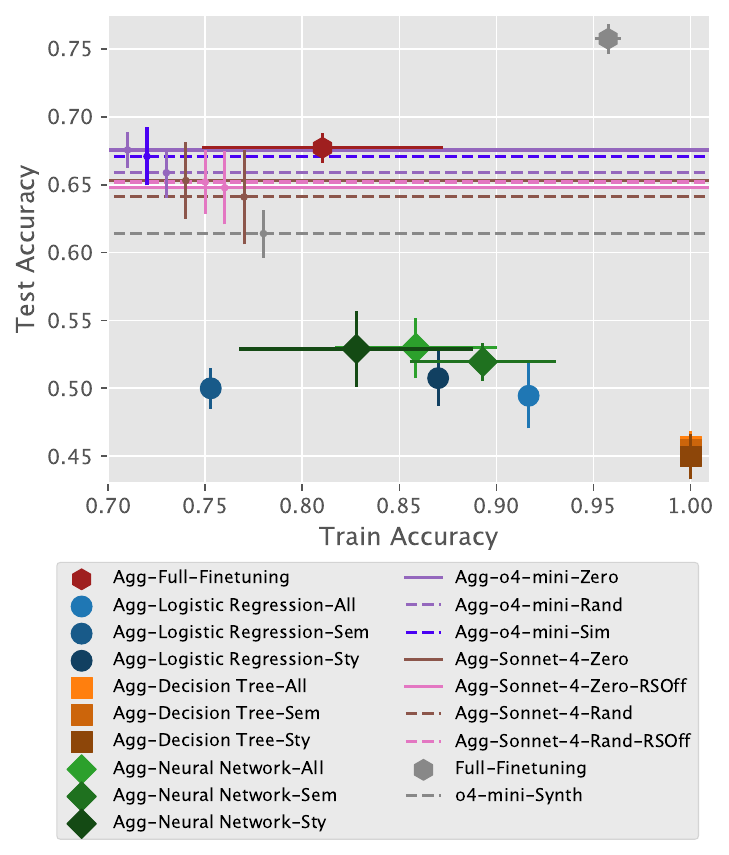}
    \caption{RQ3 results on aggregated preference modeling. \texttt{Zero} indicates that LLMs are prompted in zero-shot. Other label descriptions are provided in Figure~\ref{fig:rq2}. We included the best results for model training (\texttt{Full\allowbreak-finetuning}) and LLM prompting (\texttt{o4\allowbreak-mini\allowbreak-Synth}) from Figure~\ref{fig:rq2} for comparison.}
    \Description{This figure presents results for aggregated preference modeling comparing various model configurations against language model baselines. The plot displays test accuracy on the vertical axis ranging from 0.45 to 0.75, and train accuracy on the horizontal axis from 0.70 to 1.00, with error bars indicating confidence intervals. The plot includes multiple horizontal reference lines representing language model performance. At the top, a solid purple line at approximately 0.68 marks the Agg-o4-mini-Zero performance, with a dashed blue line just below at 0.67 for Agg-o4-mini-Sim. A dashed purple line followed, being around 0.66, which is for Agg-o4-mini-Rand. Brown solid and pink dashed, and pink solid lines appear around 0.65 representing Agg-Sonnet-4-Zero, Agg-Sonnet-4-Rand-RSoff, and Agg-Sonnet-4-Zero-RSoff, with a corresponding brown dashed line at 0.64 for Agg-Sonnet-4-Rand. For comparison, o4-mini-Synth performance is also shown, as a gray dashed line at 0.615. The plot includes Full-Finetuning also as a comparison, shown as a gray hexagon at the top right, achieving approximately 0.76 test accuracy with 0.96 train accuracy. Agg-Full-Finetuning as a red circle at about 0.68 test accuracy with 0.81 train accuracy. Aggregated versions of Neural Network models, displayed as green diamonds, cluster around 0.52-0.53 test accuracy with train accuracies between 0.83 and 0.90. Aggregated versions of logistic Regression models, shown as blue circles, perform poorly at around 0.50 test accuracy despite achieving train accuracies between 0.75 and 0.93. Aggregated Decision Tree models, represented by orange and brown squares at the bottom right, show the worst performance at approximately 0.46 test accuracy despite perfect or near-perfect train accuracy of 1.00.}
    \label{fig:rq3}
\end{figure}

Figure~\ref{fig:rq3} presents the analysis results. We found that \textbf{\texttt{Agg-Full-Finetuning} and \texttt{Agg-o4-mini-Zero} had the highest test accuracy}. However, \textbf{their test accuracies were lower than the best personalized models (\texttt{Full-Finetuning})}. This result implies that modeling aggregated preferences is a more difficult task, possibly due to the low agreement between annotators. Moreover, we found that model training approaches other than \texttt{Agg\allowbreak-Full\allowbreak-Finetuning} had low test accuracies, being lower than LLM prompting ones. For LLM prompting, few-shot examples did not help but even decreased the test accuracy. With few-shot examples, similarity-based sampling performed better than random sampling. Moreover, \textbf{all LLM prompting approaches for aggregated preference prediction had higher test accuracies than the best performing one for personal preference prediction (\texttt{o4-mini-Synth}).} It might be because they are trained on aggregated preferences---hence, they perform better at inferring aggregated preferences than predicting individual user preferences.

\subsection{RQ4: Can we leverage stated preferences to model personal revealed preferences?}
\label{sec:rq4}

\subsubsection{Motivation}
As we collected stated preferences, we were curious if they could help predict personal preferences. Hence, we analyzed approaches that consider stated preferences.

\subsubsection{Analysis Method}
We focused on analyzing modeling approaches that consider both 1) revealed and 2) stated preferences. Hence, all modeling approaches we examined in this analysis made predictions across annotators, not focusing on a single annotator. Due to this, we prepend \texttt{Cross} to the condition names to differentiate them from those from RQ2 (Section~\ref{sec:rq2}). Note that when considering the annotator profile, along with stated preferences, we also considered demographics, which are in Figure~\ref{fig:demographic}.

For the model that we trained, the main analysis-wise difference from RQ2 was in the data split. We had two types of test datasets: 1) cross-annotator test sets and 2) within-annotator test sets. First, we chose 10\% of annotators and considered their annotations as a cross-annotator test set. Then, for the remaining annotators, we took 10\% of annotations from each annotator and considered the union of them as within-annotator test sets. By splitting test sets in this way, we could measure whether the trained models generalize to unseen annotators and unseen instances from already seen annotators. Hence, we calculated two test accuracy metrics: 1) cross-annotator test accuracy and 2) within-annotator test accuracy. Note that while we ran the full 10-fold validation for within-annotator test sets, we ran only five folds for cross-annotator test sets, resulting in a total of 50 folds. 
For LLM prompting approaches, we did not consider cross-annotator test sets as we prompted models to consider only within-annotator examples. 

We adopted the same technical approaches as RQ2, but they required modifications to consider both stated and revealed preference data. Moreover, we considered one additional condition, which predicts weights for logistic regression models from the stated preference inputs (\texttt{Cross-LR-Weight}).

\paragraph{Finetuned ModernBERT-large (\texttt{Cross\allowbreak-Full\allowbreak-Finetuning})} 
To incorporate stated preferences in finetuning (\texttt{Cross-} in Figure~\ref{fig:modeling_approaches}a), inspired by previous work~\cite{orlikowski2025beyond}, we took the approach of prepending stated preferences to both winning and losing instances. By training with stated preferences, we expect the model to learn to differentiate preferences between annotators. Specifically, we appended numerical and ordinal answers (see Table~\ref{tab:stated_preference_questions}) as numerical values while listing multiple selections as a list of selected category names. We also appended optional open-ended responses for multiple selection questions. Note that, as we are finetuning weights, if the input format is consistent, it is okay to omit questions. Responses to each question were separated with \texttt{[SEP]} tokens. Winning and losing texts were appended after this stated preference input, with \texttt{[SEP]} tokens used as separators. As we are calculating two test accuracy metrics, we picked those that have the best average metric.

\paragraph{Logistic regression, Decision tree, and Neural Network (\texttt{Cross\allowbreak-\{Logistic Regression, Decision Tree, Neural Network\}\allowbreak-\{All\}})} 
For the models that are trained on embedded texts, we incorporated stated preferences by turning them into vectors (\texttt{Cross-} in Figure~\ref{fig:modeling_approaches}b). Specifically, we transformed numerical and ordinal values into floats with normalization so that the maximum value would be one. For the ordinal values, we assumed an equal distance between orders. We turned multiple selection values into one-hot values, where the existence of the option is marked as one. With this process, we transformed the state preferences into vectors with 93 dimensions. When training models, we appended these stated preference vectors to the embeddings of the input texts. For neural network models, similar to ModernBERT-large full-finetuning, we picked the results with the best means of two test accuracy metrics. We only considered cases where we used the concatenation of semantic and style embeddings, as generally combining them seems to show better performance in previous sections.

\paragraph{LLM prompting (\texttt{Cross\allowbreak-o4\allowbreak-mini\allowbreak\{, -Rand, -Sim, -Synth\}} and \texttt{Cross\allowbreak-Sonnet\allowbreak-4\{, -Rand\}\{, -RSOff\}})}
For approaches that prompt LLMs, we added the stated preference information into the prompts. In these prompts, we also listed optional open-ended responses for multiple selection questions. Note that there can be conditions without few-shot examples, but only with state preferences. Please refer to Appendix~\ref{app:usedprompts} for details.

\paragraph{Predicting logistic regression weights from stated preferences (\texttt{Cross\allowbreak-LR\allowbreak-Weight})}
We examined a condition where we use stated preferences to predict a specific annotator's preference annotation model (Figure~\ref{fig:modeling_approaches}c). 
As the format of the inferred preference annotation model, we used logistic regression over embedded texts, as the model showed acceptable performance when trained on individual annotators (Section~\ref{sec:rq2}). Moreover, we observe that these weights could be used as interpretable vectors for the user's preferences~\cite{kim2018interpretability}. Specifically, given the stated preference vector for an annotator ($u^i$ for the annotator $i$, the same format as used for embedding-based models), we predicted the weights ($W_{\text{predicted}}^i$) and biases ($b_{\text{predicted}}^i$) for the annotator with the N-layer neural network ($f$):
\begin{equation}
(W_{\text{predicted}}^i, b_{\text{predicted}}^i) = f(u^i)
\end{equation}
Note that we had the base weights and biases as learnable parameters, and the neural network predicted only the residual term. The final weights ($W_{\text{final}}^i$) and biases ($b_{\text{final}}^i$) are calculated as follows:
\begin{equation}
W_{\text{final}}^i = W_{\text{base}} + 0.1W_{\text{predicted}}^i
\end{equation}
\begin{equation}
b_{\text{final}}^i = b_{\text{base}} + 0.1b_{\text{predicted}}^i
\end{equation}
Then, final weights and biases are used for the prediction, given a pair of texts as input. In the experiment, we used the text embeddings that concatenated both semantic and style embeddings. We used 2-layer neural networks with a hidden layer size of 1024. The initial learning rate was 1e-6 with linear decay, and the batch size was 16. We trained models with a maximum of 100 epochs and 20 epochs of early stopping threshold. We took the average of cross-annotator and within-annotator test accuracies as the metric for early stopping.

\subsubsection{Results}

\begin{table*}[]
\caption{RQ4 results on whether we could leverage stated preferences to model personal reading preferences.}
\adaptivetablesize
\setlength{\tabcolsep}{4px}
\centering
\begin{tabular}{c|c|cc|c}
\thickhline
Condition & Cross-Annotator Test Acc & Within-Annotator Test Acc & Diff to Non-Cross Counterpart & Train Acc \\ \thickhline
\texttt{Cross-Full-Finetuning} & $0.522 \pm 0.010$ & $0.544 \pm 0.006$ & {\color[HTML]{9e082b} $-0.213$} & $0.549 \pm 0.007$ \\
\texttt{Cross-Logistic Regression-All} & $0.522 \pm 0.016$ & $0.551 \pm 0.007$ & {\color[HTML]{9e082b} $-0.073$} & $0.747 \pm 0.001$ \\
\texttt{Cross-Decision Tree-All} & $0.499 \pm 0.007$ & $0.507 \pm 0.004$ & {\color[HTML]{9e082b} $-0.038$} & $1.000 \pm 0.000$ \\
\texttt{Cross-Neural Network-All} & $0.591 \pm 0.007$ & $0.589 \pm 0.004$ & {\color[HTML]{9e082b} $-0.123$} & $0.977 \pm 0.045$ \\
\texttt{Cross-o4-mini} & - & $\underline{0.614 \pm 0.014}$ & - & - \\
\texttt{Cross-o4-mini-Rand} & - & $0.592 \pm 0.014$ & {\color[HTML]{085217} $+0.027$} & - \\
\texttt{Cross-o4-mini-Sim} & - & $0.592 \pm 0.014$ & {\color[HTML]{085217} $+0.037$} & - \\
\texttt{Cross-o4-mini-Synth} & - & $0.613 \pm 0.018$ & {\color[HTML]{9e082b} $-0.001$} & - \\
\texttt{Cross-Sonnet-4} & - & $0.598 \pm 0.014$ & - & - \\
\texttt{Cross-Sonnet-4-Rand} & - & $0.578 \pm 0.014$ & {\color[HTML]{085217} $+0.025$}  & - \\
\texttt{Cross-Sonnet-4-RSOff} & - & $0.595 \pm 0.014$ & - & - \\
\texttt{Cross-Sonnet-4-Rand-RSOff} & - & $0.573 \pm 0.014$ & {\color[HTML]{085217} $+0.013$} & - \\
\texttt{Cross-LR-Weight} & $\mathbf{0.634 \pm 0.008}$ & $\mathbf{0.624 \pm 0.006}$ & - & $0.642 \pm 0.004$ \\
{\color[HTML]{9B9B9B} \texttt{Full-Finetuning}}& {\color[HTML]{9B9B9B} -} & {\color[HTML]{9B9B9B} $0.757 \pm 0.011$} & {\color[HTML]{9B9B9B} -} & {\color[HTML]{9B9B9B} $0.958 \pm 0.007$} \\
{\color[HTML]{9B9B9B} \texttt{o4-mini-Synth}} & {\color[HTML]{9B9B9B} -} & {\color[HTML]{9B9B9B} $0.614 \pm 0.018$} & {\color[HTML]{9B9B9B} -} & {\color[HTML]{9B9B9B} -} \\ \thickhline
\end{tabular}
\label{tab:rq4}
\Description{This table presents results examining whether stated preferences can effectively model personal reading preferences, comparing cross-annotator and within-annotator test accuracies across different models and conditions. The table shows four columns of data: Cross-Annotator Test Accuracy, Within-Annotator Test Accuracy, Diff to Non-Cross Counterpart, and Train Accuracy. Cross-Full-Finetuning achieves 0.522 cross-annotator and 0.544 within-annotator test accuracy with a substantial drop of 0.213 from its non-cross counterpart. Its train accuracy was 0.549. Cross-Logistic Regression-All shows similar cross-annotator performance at 0.522 but slightly higher within-annotator accuracy at 0.551, with a smaller drop of 0.073. Its train accuracy was 0.747. Cross-Decision Tree-All performs worst at 0.499 cross-annotator accuracy and 0.507 within-annotator accuracy, despite perfect training accuracy of 1.000. The performance drop compared to the non-cross counterpart was 0.038. Cross-Neural Network-All achieves cross-annotator performance at 0.591, with 0.589 within-annotator accuracy and 0.977 train accuracy. The performance drop compared to the non-cross counterpart was 0.123. LLM prompting conditions followed, and note that these do not have cross-annotator test accuracy as they are not trained. Cross-o4-mini achieves 0.614 within-annotator accuracy, while Cross-o4-mini-Rand and Cross-o4-mini-Sim both reach 0.592, showing small positive differences of 0.027 and 0.037 respectively compared to their non-cross versions. Cross-o4-mini-Synth achieved 0.613 within-annotator accuracy, with diff to non-cross counterpart being -0.001. Cross-Sonnet-4 achieved 0.598 within-annotator accuracy. Cross-Sonnet-4-Rand achieved 0.578 within-annotator accuracy, whose performance gain compared to non-cross counterpart was 0.025. Cross-Sonnet-4-RSOff achieved within-annotator test accuracy of 0.595. Cross-Sonnet-4-Rand-RSOff achieved the within-annotator test accuracy of 0.573, with diff to non-cross counterpart being 0.013. Cross-LR-Weight stands out with the highest performance at 0.634 cross-annotator and 0.624 within-annotator accuracy. Its test accuracy was 0.642. For comparison, standard Full-Finetuning achieves 0.757 within-annotator accuracy and 0.958 test accuracy, and o4-mini-Synth reaches the within-annotator accuracy of 0.614.}
\end{table*}

Table~\ref{tab:rq4} summarizes the results. In terms of within-annotator test accuracy, we could compare models from this analysis to those in RQ2 (trained on a single annotator), as both analyses' test sets contain unseen instances from the same annotators as the training data.   
We found that \textbf{models trained with stated preferences performed worse than those trained for a single annotator} (``Diff to Non-Cross Counterpart'' column in Table~\ref{tab:rq4}). This result signals that training supervised classification-style models that are aware of annotator differences is more difficult than modeling a single person's preferences. 
\textbf{Among LLM prompting approaches, only using stated preferences achieved the highest within-annotator accuracy}. Moreover, {adding the stated preferences to the prompt could boost within-annotator test accuracy} (except for \texttt{Cross-o4-mini-Synth}). In these cases, the stated preferences could provide more information about the annotator than examples, so that LLM could maximally leverage pretrained knowledge. Moreover, using the stated preferences together with examples or synthesized user profiles could have confused LLMs, compared to only using the stated preferences. \textbf{Among all approaches, \texttt{Cross-LR-Weight}, which predicts logistic regression weights out of stated preferences, achieved the best cross-annotator and within-annotator test accuracy} (while train accuracy was relatively low). 
Comparing this approach with other training approaches that more comprehensively combine stated and revealed preferences, this result implies that models struggled to learn the complex relationships between the stated and revealed preferences. 
Note that most of the approaches examined in RQ4 had accuracy close to 50\%, which is a random chance.
Considering these results, we conclude that, while stated preference has some information relevant to revealed preferences, at least with our dataset, \textbf{inferring a specific annotator's revealed preferences from stated preference is challenging, having no benefit compared to modeling approaches that focus on a single annotator}. 

\subsection{RQ5. How do people vary in their reading preferences?}
\label{sec:rq5}
\subsubsection{Motivation}
In previous sections, we explored the existence of personal tastes in creative writings and whether modeling revealed preferences is feasible. Here, we try to understand how people's preferences differ from each other. 

\subsubsection{Analysis Method}

Qualitatively analyzing annotator preferences from revealed preferences is challenging because, per annotator, there are 100 pairs of texts with the annotator's preference. Manually reading through 100 pairs for all 60 annotators is practically not feasible. To overcome this challenge, we adopted an LLM-driven approach to analyze ``how'' annotator preferences vary, which builds upon a previous work~\cite{lam2024concept}. 

\begin{figure*}
    \centering
    \includegraphics[width=\textwidth]{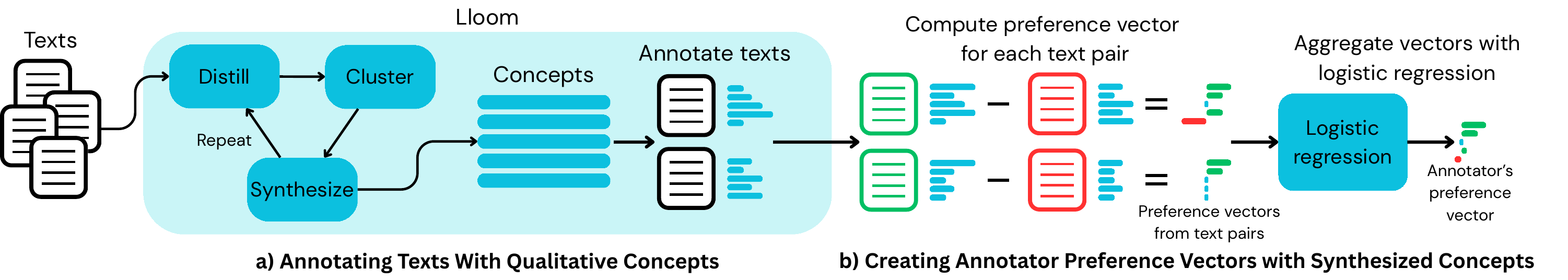}
    \caption{Parts of the analysis pipeline for RQ5.}
    \label{fig:rq5_analysis_overview}
    \Description{This figure illustrates the analysis pipeline for Research Question 5, which examines how to capture human reading preferences from the dataset. The process flows from left to right through several distinct stages. The pipeline begins with a collection of text documents on the left side. These texts enter a blue-shaded section labeled "Lloom", which has an annotation, a) Annotating Texts With Qualitative Concepts, at the bottom. Texts undergo three key operations in this section: Distill, Cluster, and Synthesize. The Synthesize step connects to horizontal blue bars representing different concept categories. This three-step process repeats iteratively until getting the refined concepts. Moving rightward, the generated concepts are used to annotate the original texts. The diagram shows document icons with colored annotation marks in blue, indicating how each text receives concept-based annotations. The next stage, labeled b) Creating Annotator Preference Vectors with Synthesized Concepts, involves computing preference vectors for text pairs. The diagram displays this through a visual equation where a green document minus a red document equals a preference vector shown in a bar chart, representing how preferences are calculated from pairwise comparisons. Multiple such preference vectors are computed from different text pairs, shown as a collection of vector representations. In the final stage on the right, these individual preference vectors are aggregated using logistic regression, depicted by a blue rounded rectangle. The logistic regression model combines all the preference vectors from text pairs to produce a single annotator's preference vector, shown as the final output vector.}
\end{figure*}

\paragraph{Annotating Texts With Qualitative Concepts (Figure~\ref{fig:rq5_analysis_overview}a)}
First, we used Lloom~\cite{lam2024concept}, an LLM-powered analysis approach, to extract high-level concepts from text snippets and annotate the emphasis of those concepts for each text. 
To extract concepts, Lloom 1) \textit{distills} the text inputs with LLMs so that they can be processed by LLMs in a reasonable length in the later part of LLM functions, 2) \textit{clusters} texts into conceptually relevant groups, 3) \textit{synthesizes} concepts out of the clusters, and 4) \textit{repeats} the aforementioned steps until we get non-overlapping, distinguishable high-level concepts. Then, with the extracted concepts, Lloom annotates the weights of concepts for text snippets in a 5-level Likert scale. We considered the 5-level scale as uniformly intervaled values between 0 and 1. 

\paragraph{Creating Annotator Preference Vectors with Synthesized Concepts (Figure~\ref{fig:rq5_analysis_overview}b)}
After running Lloom, the annotated emphasis values for different concepts could comprise a vector ($v$) that explains the overall characteristics of the text. Then, for an annotator's revealed preference over a pair of texts, we can calculate the annotator's interpretable preference vector. Assuming there is an annotator $i$ and a pair $j$, the preference vector ($p_i^j$) would be:
\begin{equation}
  p_i^j = v^j_\text{chosen} - v^j_\text{rejected}  
\end{equation}
Then, we can aggregate $p_i^j$ across all $j$s to get the vector that can explain the annotator's preference. For the aggregation, we trained logistic regression models over the preference vectors. Then, we took the coefficient of the logistic regression model~\cite{kim2018interpretability} as the aggregated vector for the annotator's preference. 

\paragraph{Clustering The Preference Vectors Of Similar Users}
As presenting all annotator preference vectors would be overloading, we instead ran clustering over all annotators' preference vectors and report the preference vectors of each cluster (i.e., centroid). We used hierarchical clustering and chose the number of clusters by locating the knee, or the maximum curvature of the plot, in how the distance metric decreases as we increase the number of clusters. We adopted polynomial interpolation when calculating the knee. After identifying annotator clusters, we conducted exploratory analyses on how different clusters vary in aspects other than their preference vector values (e.g., which cluster prefers LLM-generated texts).

\paragraph{Creating Aggregated Preference Vectors}
We also computed the vectors for aggregated preferences to learn on which characteristics all annotators ``agreed'' to prefer. For each pair, as we had preference annotations from three annotators, we aggregated annotations via majority voting. Then, with the aggregated labels, we calculated vectors for all text pairs and trained a logistic regression model over them to use its coefficient~\cite{kim2018interpretability} as an aggregated preference vector.

\paragraph{Confirming The Validity Of Preference Vectors}
As we were aware that there could be limitations in LLM-driven qualitative analysis, we also conducted a technical evaluation of this approach. We evaluated the quality of the preference vector ($p_i^j$) with human evaluators. Specifically, we showed evaluators the preference profile from $p_i^j$ along with the winning and losing texts, but without specifying which one is the winning one. Then, we asked them which text should be picked as the winning one if we follow the preference profile. We ran this evaluation over 70 randomly sampled preference annotations by asking Prolific workers in the USA and the UK, with the acceptance rate higher than 95\%. We asked each worker to annotate 10 pairs, while paying them \textsterling3 per participant (about \textsterling9 hourly payment rate). We collected three evaluations per text pair to aggregate them with majority voting. We hired 21 workers in total. Note that we had an attention check question to filter out low-quality results. The evaluation interface was deployed with Potato (see Appendix~\ref{app:techeval_interface} for the screenshot).

\begin{table*}[]
\caption{Concepts extracted from the text snippet corpus with Lloom.}
\small
\centering
\begin{tabular}{p{0.23\textwidth}p{0.6\textwidth}p{0.1\textwidth}}
\thickhline
Concept                               & Description                                                                                                                                  & Merged                    \\ \thickhline
Family and Relationships              & Does the text focus on family dynamics, interpersonal relationships, or generational conflict as a central theme or narrative driver?        &                           \\ \hline
Loss and Grief                        & Does the text explore themes of loss, grief, mourning, or emotional vulnerability related to separation or death?                            &                           \\ \hline
Conflict and Survival                 & Does the text depict physical, psychological, or moral conflict, often in the context of survival, danger, or adversity?                     &                           \\ \hline
Social Hierarchy and Class            & Does the text address issues of social class, hierarchy, reputation, or societal expectations as a key element of its content or conflict?   &                           \\ \hline
Suspense and Tension                  & Does the text create suspense, tension, or a sense of anticipation through tone, pacing, or narrative devices?                               &                           \\ \hline
Identity and Transformation  & Does the text center on questions of personal identity, self-discovery, or significant transformation (emotional, physical, or existential)? & Introspective Depth       \\ \hline
Blending Genres or Realities          & Does the text blend multiple genres (such as fantasy and realism) or blur the boundaries between reality and the fantastical/surreal?        &                           \\ \hline
Memory and Time                       & Does the text employ motifs of memory, nostalgia, time, or the passage of life as a literary device or thematic focus?                       &                           \\ \hline
Dialogue-Driven Characterization      & Does the text use dialogue as a primary means to reveal character traits, relationships, or advance the plot?                                & Dialogue Characterization \\ \hline
Genre Conventions                     & Does the text clearly utilize recognizable conventions, tropes, or stylistic features of a specific literary genre?                          &                           \\ \hline
Metaphor and Personification & Does the text prominently feature metaphors and/or personification as literary devices to convey meaning or emotion?                         &                           \\ \hline
Repetition and Fragmentation          & Does the writing style employ repetition or fragmented sentence structures for emphasis, rhythm, or to reflect emotional states?             &                           \\ \hline
Vivid Sensory Imagery                 & Does the text employ vivid sensory or descriptive imagery to create a strong sense of atmosphere, setting, or physical experience?           & Sensory Atmosphere        \\ \thickhline
\end{tabular}
\label{tab:concepts}
\Description{This table presents the concepts extracted from the text snippet corpus using Lloom. The table contains three columns showing the concept name, its description formulated as a question for annotation, and merged category labels where applicable. The table lists thirteen distinct concepts that capture various literary dimensions. Family and Relationships examines whether texts focus on family dynamics, interpersonal relationships, or generational conflict as central themes. Loss and Grief identifies texts exploring mourning, emotional vulnerability, and themes of separation or death. Conflict and Survival captures physical, psychological, or moral conflict within contexts of survival and adversity. Social Hierarchy and Class addresses issues of social stratification, reputation, and societal expectations as key narrative elements. Suspense and Tension identifies texts that create anticipation through tone, pacing, or narrative devices. Identity and Transformation centers on personal identity questions, self-discovery, and significant transformations, with which Introspective Depth has been merged. Blending Genres or Realities captures texts that mix different genres or blur boundaries between reality and the fantastical. Memory and Time examines the use of memory, nostalgia, and temporal passage as literary devices. Dialogue-Driven Characterization identifies texts using dialogue as the primary means for revealing character traits and advancing plot, with which Dialogue Characterization has been merged. Genre Conventions captures texts that clearly utilize recognizable features of specific literary genres. Metaphor and Personification focuses on texts prominently featuring these literary devices for conveying meaning. Repetition and Fragmentation examines writing styles employing repeated or fragmented sentence structures for emphasis or emotional effect. Finally, Vivid Sensory Imagery identifies texts using descriptive sensory language to create atmosphere and physical experience, with which Sensory Atmosphere has been merged.}
\end{table*}

\subsubsection{Results}
\paragraph{Concepts Obtained from Lloom}
Table~\ref{tab:concepts} shows 13 high-level concepts extracted by Lloom. Note that while it extracted more concepts than presented in Table~\ref{tab:concepts}, we merged those that are semantically similar to each other (``Merged'' in the table). We used these concepts to annotate text snippets and compute the preference vectors.

\paragraph{Confirming the Validity of Preference Vectors}
Before presenting results on preference vectors, we first show results on the technical evaluation over the analysis pipeline. When we provided evaluators with a preference vector and a pair of text snippets, they could correctly identify the preferred text with an accuracy of 90.0\%. This result indicates that each $p_i^j$, the preference vector, conveys accurate information about the preferred textual characteristic when comparing two texts in the pair. Note that we aggregated multiple of these preference vectors to compute each annotator's preference vector---while some information might be lost with the aggregation, we believe that the aggregated preference vector would provide an overview of the annotator's preference.

\begin{figure}[t]
    \centering
    \includegraphics[width=0.478\textwidth]{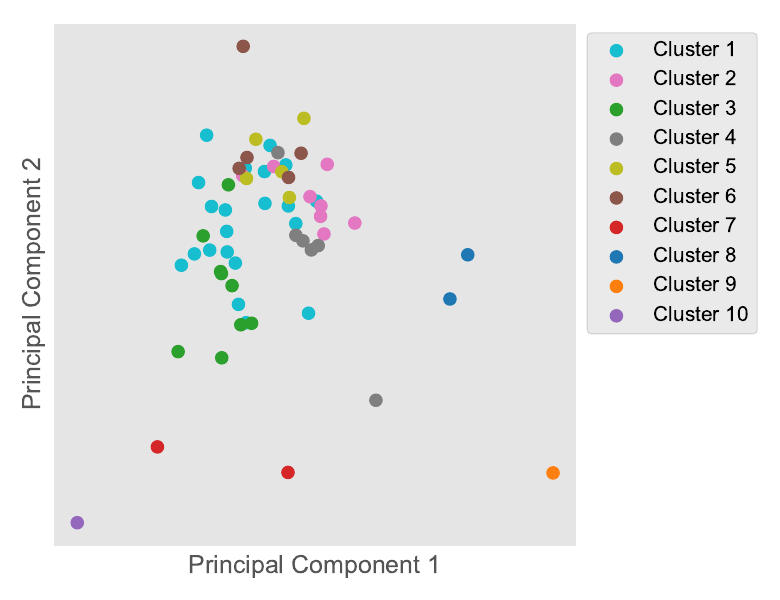}
    \caption{Annotator preference vectors with PCA on two components. We could retrieve 10 clusters.}
    \label{fig:rq5_clusters}
    \Description{This figure displays a scatter plot showing annotator preference vectors after Principal Component Analysis reduction to two dimensions, revealing ten distinct clusters of reading preferences. The plot uses Principal Component 1 on the horizontal axis and Principal Component 2 on the vertical axis, with each colored dot representing an individual annotator's preference vector. The visualization shows clear clustering patterns among the annotators. The largest concentration of points appears in the upper-left quadrant, where Clusters 1, 2, 3, and 5 intermingle, with cyan, pink, green, and yellow dots forming a dense cloud. This suggests these annotators share similar reading preferences despite being assigned to different clusters. Cluster 1 in cyan dominates this region with approximately 15-20 points, while Clusters 3 in green and 5 in yellow show substantial overlap in this same area. Several clusters appear more isolated, indicating distinct preference patterns. Cluster 7 in red appears in the lower-left area with just two points, suggesting a unique preference profile. Cluster 8 in blue has three points in the center-right and far-right portions of the plot, showing separation from the main group. Cluster 9 in orange has a single point in the lower-right corner, representing the most isolated preference pattern. Cluster 10 in purple appears in the far lower-left with a single point, also showing distinct preferences. Clusters 4 and 6 are represented by gray and brown dots respectively, with Cluster 4 having several points scattered in the central area and one outlier in the center-right, while Cluster 6 shows four brown points in the upper portion of the plot, with one notably separated from all other clusters at the top.}
\end{figure}

\begin{figure*}[t]
    \centering
    \includegraphics[width=\textwidth]{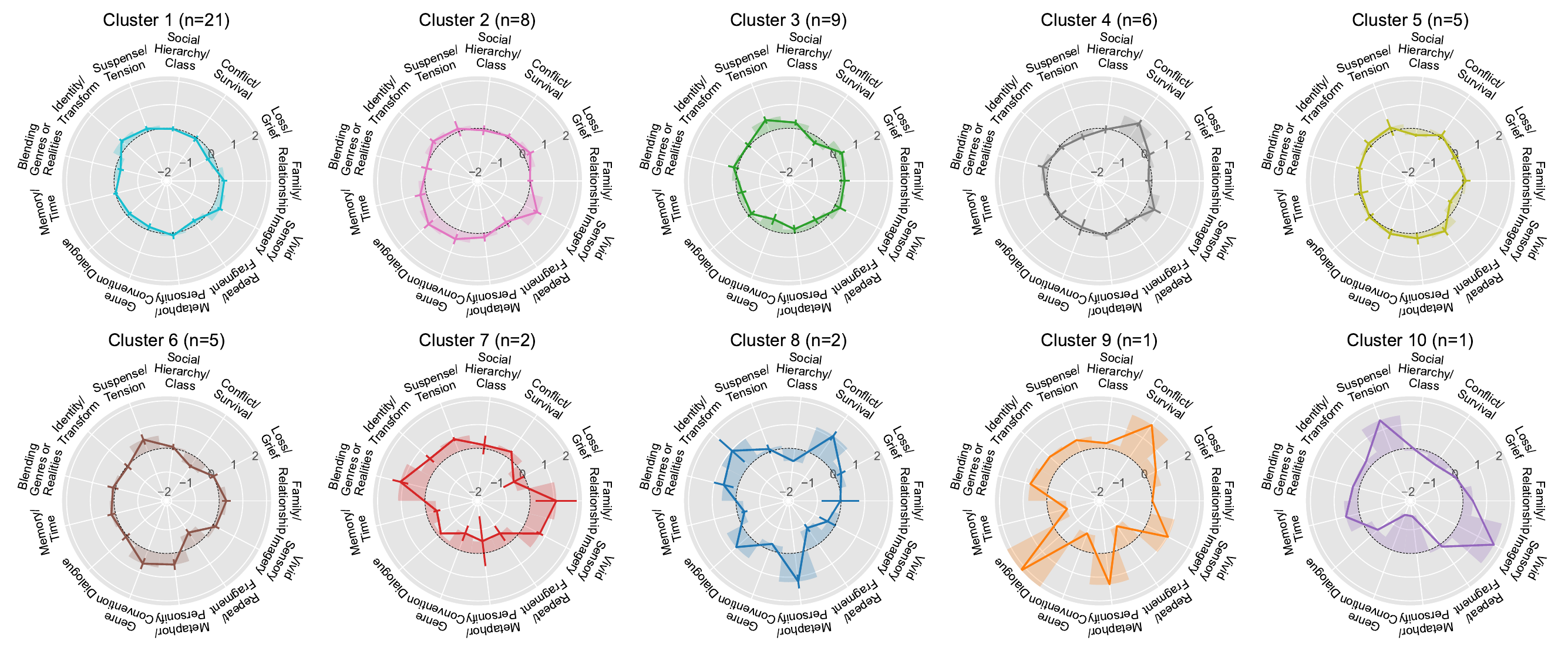}
    \caption{Annotator preference vector profiles for each cluster from Figure~\ref{fig:rq5_clusters}.}
    \label{fig:rq5_profiles}
    \Description{This figure presents radar plots showing preference vector profiles for each of the ten clusters identified in Figure 7, with each cluster's sample size indicated in parentheses. The plots display preferences across thirteen literary concepts arranged radially, with values ranging from negative 2 to positive 1 on the radial axis. Cluster 1, the largest group with 21 annotators shown in cyan, displays a relatively balanced profile with slight positive preferences for family/relationship, identity/transform, and vivid sensory image. It has negative preferences for concepts like Loss/Grief and Blending Genres or Realities. Cluster 2 with 8 annotators in pink shows a similar balanced pattern but positive preferences for Loss/Grief, Identity/Transform, Memory/Time, Dialogue, Genre Convention, Metaphor/Personify, and Vivid Sensory Image. Cluster 3 containing 9 annotators in green exhibits moderate negative preferences across many dimensions except for Loss/Grief, Social Hierarchy/Class, Suspense/Tension, Blending Genres or Realities, Family/Relationship, andVivid Sensory Image. Among the negative ones, Genre Convention got high emphasis. Cluster 4 with 6 annotators in gray shows hovering close to neutral across all concepts except for a bit high positive preferences for Conflict/Survival and Vivid Sensory Imagery. Cluster 5 with 5 annotators in yellow displays slight positive for Repeat/Fragment and negative for Social Hierarchy/Class and Vivid Sensory Imagery. Cluster 6, also with 5 annotators shown in brown, shows a bit intense taste, with positives on Family/Relationship, Suspense/Tension, Genre Convention, Metaphor/Personify, and negatives on Conflict/Survival, Repeat/Fragment. Cluster 7 to 10 has even intense preferences. Cluster 7, containing only 2 annotators in red shows more pronounced variation, with stronger positive preferences for Family/Relationship, Suspense/Tension, Identity/Transform, Blending Genres or Realities, and Vivid Sensory Imagery, while being negative on Loss/Grief, Memory/Time, Genre Convention, Metaphor/Personify, and Repeat/Fragment. Cluster 8, also with 2 annotators in blue, displays positive preferences for Conflict/Survival, Identity/Transform, Blending Genres or Realities, Dialogue, Metaphor/Personify while having negative preferences for Social Hierarchy/Class, Repeat/Fragment, and Vivid Sensory Imagery. Cluster 9 with a single annotator in orange exhibits the most extreme profile, with sharp spikes outward indicating strong positive preferences for several concepts including Conflict Survival, Blending Genres or Realities, Dialogue, Metaphor/Personify, and Vivid Sensory Imagery, while showing deep negative preferences for Memory/Time, Genre Convention, and Repeat/Fragment, creating a distinctive star-like pattern. Cluster 10, also representing a single annotator in purple, shows positive preference for Suspense/Tension, Memory/Time, and Vivid Sensory Imagery, while being negative on Genre Convention and Metaphor/Personify.}
\end{figure*}

\paragraph{Analyzing Clusters of User Preference Vectors}
Figure~\ref{fig:rq5_clusters} shows how annotator preference vectors are distributed and clustered when the dimensions are reduced to two with PCA. Figure~\ref{fig:rq5_profiles} presents how each cluster varies in terms of concepts they prefer or do not prefer. We found that the majority of annotators tend not to have too strong tastes. That is, clusters in the upper row of Figure~\ref{fig:rq5_profiles} are large in size and do not have vector values deviating too much from 0. They were also somehow more closely clustered in Figure~\ref{fig:rq5_clusters}. However, they still vary in their ``preference directions.'' For example, while \texttt{Cluster 1} does not necessarily prefer dialogues, \texttt{Cluster 2} has a clear preference for dialogue-driven characterization. 

Clusters in the bottom row of Figure~\ref{fig:rq5_profiles} were smaller in size while having more intensive weights in their preference vectors. They were also more outlying in Figure~\ref{fig:rq5_clusters}. Moreover, their direction of preference did not converge. For instance, \texttt{Cluster 9} (which is one annotator) had a very strong preference for dialogue elements, not at a comparable level to any other clusters. Overall, \textbf{our analysis shows that annotator preferences diverge, both in terms of their directions and intensities in tastes}.

\begin{table*}[t]
\caption{The top 5 most agreed demographics and stated preferences questions for each cluster. We did not list clusters with fewer than three annotators.}
\small
\centering
\begin{tabular}{p{0.07\textwidth}|p{0.16\textwidth}p{0.16\textwidth}p{0.16\textwidth}p{0.16\textwidth}p{0.16\textwidth}}
\thickhline
Cluster \#       & Top 1                                                       & Top 2                                           & Top 3                                                       & Top 4                                                           & Top 5                                                                         \\ \thickhline
\texttt{Cluster 1} (n=21) & Why you read: For relaxation / stress release (n=19)        & Read textual content: Every day (n=17)          & Why you read: Expand my world view (n=16)                   & Why you read: Learn about topics that interest me (n=16)        & Why you read: Learn about the world through other people's experiences (n=15) \\ \hline
\texttt{Cluster 2 }(n=8)  & Why you read: Stimulate my imagination and creativity (n=7) & Why you read: Expand my world view (n=7)        & Why you read: For relaxation / stress release (n=7)         & English reading skill: Native/Near-native (n=7)                 & Why you read: Learn about topics that interest me (n=6)                       \\ \hline
\texttt{Cluster 3} (n=9)  & Preferred genre (fiction): Crime / mystery / thriller (n=9) & Why you read: Expand my world view (n=9)        & Why you read: Learn about topics that interest me (n=9)     & Preferred genre (fiction): Contemporary / general fiction (n=8) & Why you read: Drama of good stories / watch a good plot unfold (n=8)          \\ \hline
\texttt{Cluster 4} (n=6)  & English reading skill: Native/Near-native (n=6)             & Watch videos: Every day (n=5)                   & Why you read: Stimulate my imagination and creativity (n=5) & Why you read: Improve my analytical / critical thinking (n=5)   & Why you read: For relaxation / stress release (n=5)                           \\ \hline
\texttt{Cluster 5} (n=5)  & Education: Graduate degree (n=5)                            & English reading skill: Native/Near-native (n=5) & Read textual content: At least once per week (n=5)          & Why you read: Expand my world view (n=5)                        & Preferred genre (non-fiction): Personal Development (n=4)                     \\ \hline
\texttt{Cluster 6} (n=5)  & Geolocation: North America (n=5)                            & Education: Graduate degree (n=5)                & Watch videos: Every day (n=5)                               & Why you read: Improve my analytical / critical thinking (n=5)   & Why you read: For relaxation / stress release (n=5)                          \\ \thickhline
\end{tabular}
\label{tab:rq5_cluster_top_stated}
\Description{This table reveals the top 5 most agreed-upon demographic and stated preference questions for clusters containing at least three members. Cluster 1, the largest with 21 annotators, is characterized primarily by reading for relaxation and stress release, with 19 of 21 members citing this motivation. They read textual content daily, with 17 members maintaining this habit. This cluster also reads to expand worldview and learn about topics of interest, with 16 members sharing each motivation, and learns through others' experiences, cited by 15 members. Cluster 2 containing 8 annotators shows a preference for reading that stimulates imagination and creativity, shared by 7 members. They also read to expand their worldview and for relaxation, each cited by 7 members. All 7 members who responded about English proficiency identify as Native or Near-native speakers, and 6 read to learn about interesting topics. Cluster 3 with 9 members unanimously prefers crime, mystery, and thriller fiction. All 9 members read to expand their worldview and learn about topics of interest. Eight members prefer contemporary and general fiction, and 8 also read for the drama of stories and watching plots unfold. Cluster 4 comprising 6 annotators all identify as Native or Near-native English speakers. Five members watch videos daily and read to stimulate imagination and creativity. They also read to improve analytical and critical thinking skills and for relaxation, each motivation shared by 5 members. Cluster 5 with 5 members all hold graduate degrees and identify as Native or Near-native English speakers. All 5 read textual content at least weekly and to expand their worldview. Four members prefer non-fiction focused on personal development. Cluster 6, also containing 5 annotators, consists entirely of North American residents with graduate degrees. All members watch videos daily and read both to improve analytical thinking and for relaxation.}
\end{table*}

For each derived cluster, we analyzed the top-5 most agreed-upon responses in demographic and stated preference questions (Table~\ref{tab:rq5_cluster_top_stated}).\footnote{Note that there could be ties in the top 5, and we presented the results that are more relevant to the cluster's preference vector profiles. We provide full results as supplementary material.} 
While not all, \textbf{some agreed-upon responses resonated with what is revealed in the preference vector of each cluster}. For example, all annotators of \texttt{Cluster 3} preferred crime, mystery, or thriller fictions, and they tend to have high suspension/tension values in their profile (top 1 in \texttt{Cluster 3} of Figure~\ref{fig:rq5_profiles}). This cluster also agreed highly that they read for good stories and plots, which might be the reason why they avoided genre conventions, such as mundane tropes. Similarly, \texttt{Cluster 2} agreed on valuing stimulation of imagination and creativity, which might be relevant to a high score on vivid sensory imagery in their vector profile. 
However, \textbf{not all agreed responses were highly related to preference vector profiles}, as some were frequently highly agreed across clusters (e.g., ``Expand my world view'' was highly agreed in four clusters). 

\begin{figure}
    \centering
    \includegraphics[width=0.478\textwidth]{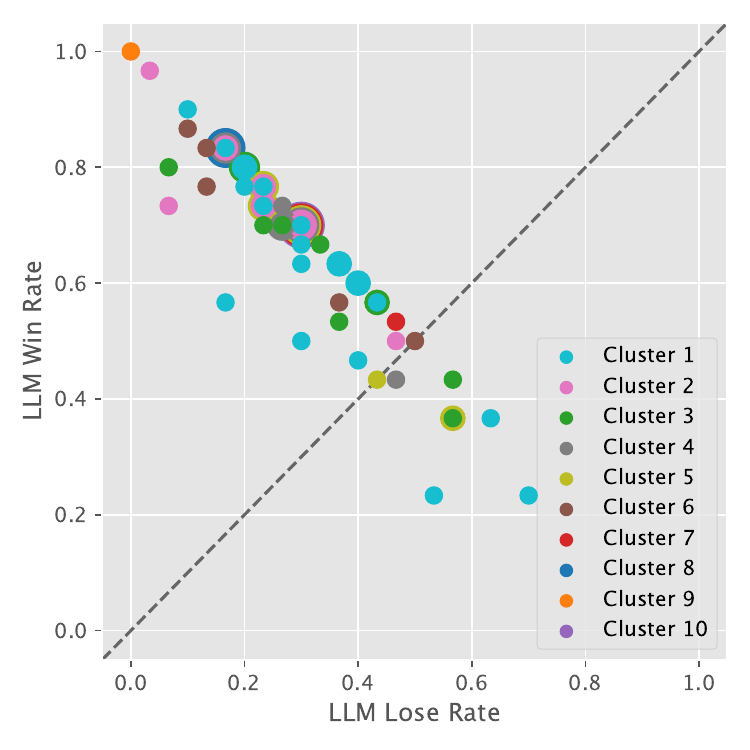}
    \caption{Win rate plot over whether annotators prefer LLM-generated texts over human-written texts. Overlapping data are expressed in varying glyph sizes.}
    \label{fig:rq5_llm_win_lose_rate}
    \Description{This figure presents a scatter plot examining annotator preferences between LLM-generated and human-written texts, with LLM Lose Rate on the horizontal axis and LLM Win Rate on the vertical axis, both ranging from 0 to 1. A diagonal dashed line represents the boundary where win and lose rates are equal, with points above indicating preference for LLM texts and below indicating preference for human texts. The plot reveals that most data points cluster above the diagonal line, indicating a general preference for LLM-generated texts across annotators. The points show considerable overlap, represented by varying glyph sizes to distinguish overlapping data, with most concentrated in the upper-left region where LLM win rates range from 0.6 to 0.9 and lose rates remain below 0.4. Cluster 1 annotators, shown in cyan, display the widest distribution across the plot, with some achieving LLM win rates as high as 0.9 while others fall closer to 0.5. Several Cluster 1 points appear in the lower right, showing higher lose rates around 0.6-0.8 with corresponding lower win rates, though still generally favoring LLM texts. Clusters 2 through 6 and 10, represented in pink, green, gray, yellow, brown, and purple, respectively, show similar patterns with most annotators achieving win rates between 0.65 and 0.85 against lose rates of 0.2 to 0.4. These clusters concentrate in the upper-middle region, demonstrating consistent moderate preference for LLM-generated content. Notable outliers include a Cluster 9 annotator in orange at the top-left corner with a perfect 1.0 win rate and near-zero lose rate, indicating exclusive preference for LLM texts. Cluster 7 in red appearㄴ as single points in the middle range, showing more balanced but still LLM-favoring preferences.}
\end{figure}

We were also curious if different clusters have different levels of preference for LLM-generated texts. Hence, for those pairs where LLM-generated texts are compared with human-written texts, we computed LLM win rates (Figure~\ref{fig:rq5_llm_win_lose_rate}). The result showed that \textbf{there are not many differences between clusters, but annotators generally seemed to prefer LLM-generated texts more than human-written ones}. Only six annotators preferred human-written texts more than LLM-generated ones.

\begin{figure}
    \centering
    \includegraphics[width=0.478\textwidth]{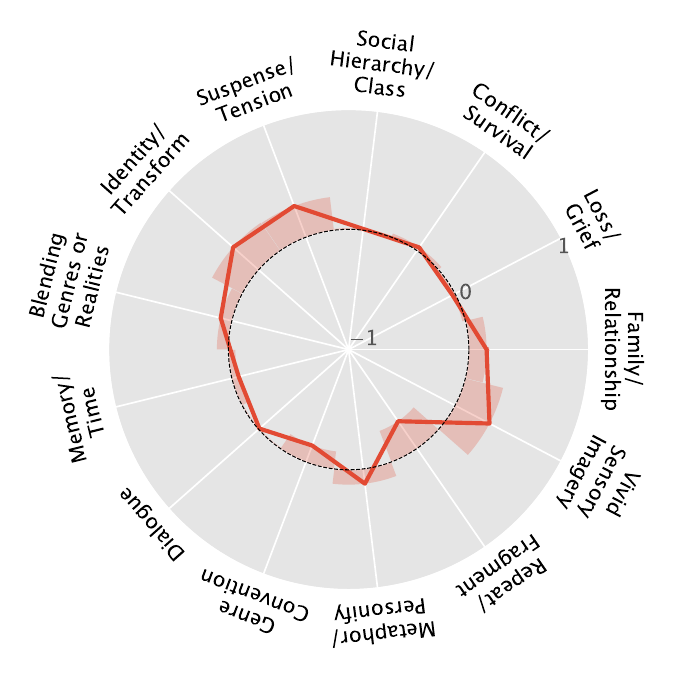}
    \caption{Aggregated preference vector profile from all annotators, aggregated with majority voting. Note that the scale is half of Figure~\ref{fig:rq5_profiles}.}
    \label{fig:rq5_global}
    \Description{This figure displays an aggregated preference vector profile combining preferences from all annotators through majority voting, presented as a radar plot with thirteen literary concepts arranged radially. The scale ranges from negative 1 to positive 1, which is half the scale used in Figure 8, providing a more focused view of the aggregate preferences. The red polygon shows the aggregated preference profile, with a black dotted line indicating the neutral zero baseline for reference. The shaded red area emphasizes the regions where preferences deviate from neutral. The profile reveals several notable patterns in collective reading preferences. Suspense/Tension, Identity/Transform, and Vivid Sensory Imagery show a strong positive preference, while Blending Genres or Realities, Family/Relationship, and Metaphor/Personify show moderate positive preference. Repeat/Fragment had high negative preference, while Genre Convention and Memory/Time followed afterward. Others, Loss/Grief, Conflict/Survival, Social Hierarchy/Class, and Dialogue were close to neutral.}
\end{figure}

\paragraph{Analyzing Aggregate Preference Vectors}
Figure~\ref{fig:rq5_global} shows the aggregated preference vector profile for all annotators. While the aggregated preference vector did not have very intense values, it still had a directionality. In aggregation, anntotors preferred suspense, identity-related topics, and vivid sensory images, while weakly preferring family-related topics, genre or reality blendings, and metaphors. Annotators in aggregation did not prefer repetition and fragmentation, while weakly avoiding genre convention.

\section{Discussion}

We discuss 1) revealed preference modeling, 2) aggregated preference modeling, 3) using stated preference data, 4) a guide for creative writing personalization, and 5) limitations and future work. 

\subsection{Modeling Personal Preferences from Revealed Preference Data}
RQ1 analysis (Section~\ref{sec:rq1}) confirms that personal taste exists for creative writing. With the analysis of RQ5 (Section~\ref{sec:rq5}), we interpret those preferences as vectors, assuming linear preference directionality. However, we acknowledge that our interpretation can have limitations, as results from RQ2 (Section~\ref{sec:rq2}) showed that non-linear modeling approaches (\texttt{Full-Finetuning} and \text{Neural Network}) were more accurate in modeling annotator preferences than linear ones (\texttt{Logistic Regression}). People's preference direction might change depending on which specific texts they are reading through, which would not be best explained with linear representations. 
LLM prompting could model personal preferences only when it synthesized a sufficient amount of revealed preference data into the user profile (\texttt{o4-mini-Synth}). However, the accuracy was only comparable to linear modeling (\texttt{Logistic Regression}), signaling that a fixed preference profile would not most accurately model personal preference.
While non-linear functions seem to help model personal preferences, one encouraging result was that, if we leverage already pretrained models, we do not need that many revealed preference samples to reach acceptable accuracy (Figure~\ref{fig:rq2_varying}). However, we did not see the accuracy plateauing with more samples, indicating that a larger dataset could add further benefits.

\subsection{Aggregated Preference}

RQ3 results (Section~\ref{sec:rq3}) show that modeling aggregated preferences is more difficult than modeling personal preferences. One possibility is that, as the aggregation is done on three specific annotators, our aggregated data could provide contradicting information depending on the sets of annotators. While finetuning the transformer encoder performed best (\texttt{Agg-Full-Finetuning}), smaller non-linear models (\texttt{Agg-Neural Network}) did not exhibit a significant performance benefit over linear ones (\texttt{Agg-Logistic Regression}), which could also be attributed to the complexity of aggregated preferences. Surprisingly, LLM prompting without any example input data (\texttt{Agg-o4-mini-Zero}) performed equivalently to \texttt{Agg-Full-Finetuning}. This indicates that some LLMs already have some amount of knowledge about ``general preferences,'' and providing a small number of examples could confuse these models. Our interpretations of generally preferred textual aspects (Figure~\ref{fig:rq5_global}) were reasonable at a high level (e.g., people generally prefer sensorily vivid imageries). However, as this interpretation is on a linear representation and linear modeling approaches are not the most accurate, this interpretation would only partially explain the aggregated preferences.

\subsection{Leveraging Stated Preference Data for Preference Modeling}

RQ4 analyses (Section~\ref{sec:rq4}) revealed that it is difficult to train a model that can infer a specific annotator's revealed preferences given their stated preferences. Three possible reasons exist: 1) questions for stated preferences were not comprehensive enough to capture how annotators would behave in revealed preference annotation, 2) the number of annotators in the dataset was not large enough to capture the full spectrum of users, or 3) stated preferences can have contradicting or unhelpful information in inferring revealed preferences. LLM prompting results, on the other hand, indicate that some stated preference information is relevant to revealed preferences. 
For example, prompting LLM only with stated preferences could achieve the second-best results for cross-annotator modeling and performed equivalently to \texttt{o4-mini-Synth}, the best performing LLM prompting approach from RQ2 analysis. 
However, combining state preference input with revealed preference data did not help in the case of LLM prompting, again indicating that stated and revealed preferences might contain contradicting information.
Interestingly, \texttt{Cross-LR-Weight}, the approach that predicts logistic regression weights out of the stated preference input, had the highest cross-annotator modeling performance.
Considering the simplicity of logistic regression and that we inferred its weight only from stated preference, the success of this unconventional model was surprising and warrants future research.
At the same time, the fact that the model does not closely combine stated and revealed preference data implies the difficulty of closely combining both types of data. Resonating with modeling results, Table~\ref{tab:rq5_cluster_top_stated} showed that stated preference can be related to interpretations of revealed preferences---but not fully explaining them.

\subsection{A Practical Guide for Eliciting Personal Preference for Creative Writing}

In practice, interactions to elicit personal preferences could take various forms, from survey questions (including open-ended ones) to binary preferences annotation (e.g., image generation personalization on Midjourney\footnote{\url{https://www.midjourney.com/personalize} and \url{https://docs.midjourney.com/hc/en-us/articles/32433330574221-Personalization}}).
With our findings, we suggest a guideline for eliciting personal preferences for creative writing, if the elicited data is to be used for modeling individual preferences with the current technologies. 
When a developer has resources to finetune a transformer encoder model per user (e.g., training time and GPUs), eliciting revealed preference data would be desirable. It is due to the high performance of such models. Collecting more samples would be more desirable, but even 15 samples would be enough for decent modeling performance. When lacking the capacity to finetune transformer-based encoders, if the developer can still run inferences on text embedding models, it would be desirable to train neural networks over embeddings of revealed preference texts. In this case, collecting about 90 samples of revealed preferences would lead to decent-performing models. Only when the model developer lacks resources to run embedding models, the developers would want to use LLM prompting with stated preference data, but without expecting high performance in modeling.

\subsection{Limitations and Future Work}

We only dealt with short text snippets. Hence, our results do not convey insights about people's preferences on aspects that only manifest in longer texts, such as narrative arcs. We also did not examine all existing modeling approaches; hence, evaluating non-examined approaches can be future work. Specifically, the performance of parameter-efficient finetuning approaches (e.g., \cite{hu2022lora}) would give us practical implications as their weight sizes are small. Moreover, collecting even larger-scale data could open new research avenues. For instance, the modeling performance could increase further with more samples per annotator. Alternatively, having data on more annotators might unlock better cross-annotator modeling. Designing more comprehensive and improved stated preference survey questions can also be future work. For the interpretation of the annotator preferences, we analyzed linear preference vectors per annotator with an LLM-based pipeline. However, it could have some limitations, such as LLMs not identifying all effective conceptual dimensions or linear vectors not explaining nuances in preferences. 
Lastly, we have not yet investigated how to leverage the data for personalized text generation or in scenarios where users interact further after the initial preference elicitation.
\section{Conclusion}
We present \dataset{}, a dataset for creative writing personalization, collected from 60 annotators with diverse reading preferences. For the dataset, from each annotator, we collected binary preference annotations over 100 pairs of short creative writing texts (revealed preferences) and self-reported reading habits and tastes (stated preferences). 
As the first step toward personalization in creative writing, we ran a series of analyses to find how existing technologies perform in modeling annotator preferences and how their preferences differ from each other. By discussing the results and a guide for personal preference modeling on creative writing, we hope our work provides a stepping stone towards personalizable creative writing technologies that can adapt to one's literary taste and provide more enjoyable AI-mediated reading experiences.  

\begin{acks}
We want to thank Midjourney for supporting this work.
\end{acks}

\bibliographystyle{ACM-Reference-Format}
\bibliography{sample-base}

\appendix
\section{Data Collection Interface}
\label{app:collection_interface}

\begin{figure}
    \centering
    \includegraphics[width=0.478\textwidth]{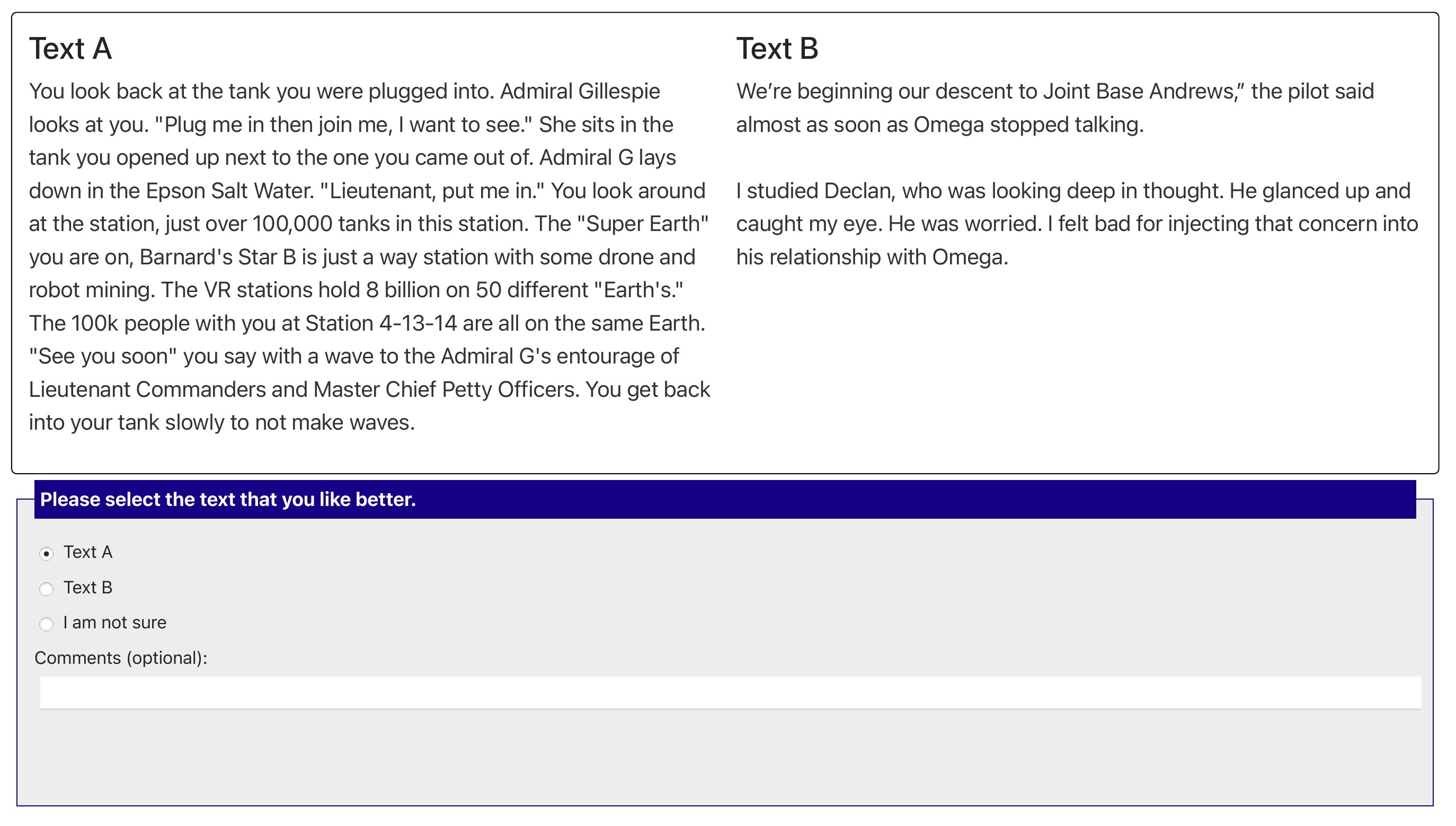}
    \caption{Data collection interface for revealed preference.}
    \label{fig:collection_interface}
    \Description{This figure displays the data collection interface used to gather revealed reading preferences from annotators. The interface presents two text passages side by side, labeled Text A and Text B, each containing a snippet of creative writing for comparison. Below the text passages, a purple header bar contains the instruction "Please select the text that you like better." For the options, there are three radio buttons, one for Text A, the second for Text B, and the last for "I am not sure." At the bottom of the interface, there is an optional comments field.}
\end{figure}
Figure~\ref{fig:collection_interface} shows the interface used for data collection.

\section{Details on Style Embedding Model}
\label{app:styleembedding}

While style embedding models exist~\cite{wegmann2022author, patel2023learning}, previous work did not focus on training models in the creative writing domain. Hence, we trained one specifically for creative writing style. For the dataset, we used Sterman et al.~\cite{sterman2020interacting}'s dataset, where the authors collected triplets of texts that consist of an anchor text, one text snippet that is more similar to the anchor, and the other that is less similar to the anchor. We finetuned ModernBERT-large~\cite{warner2025smarter} using the SentenceTransformers~\cite{reimers2020multilingual} library. We used 90\% of the dataset as a training set while the rest was a test set. Specifically, we trained the model for 10 epochs, with a batch size of 16, a learning rate of 6e-6, a linear scheduler, and a warm-up ratio of 0.1. The model was evaluated for every epoch. From 10 epochs, we picked the model with the best test accuracy, which was 0.7737. 

\section{Used Prompts}
\label{app:usedprompts}

We present the prompt used in LLM prompting conditions below. 

\begin{framed}
\noindent
\{ \texttt{if inferring an annotator's preference} \}

Your task is, for \{ \texttt{N} \} sets of tasks with two texts, to determine which of the two texts a user prefers based on their previous preferences.\\
\{ \texttt{else if inferring aggregated preference} \}

Your task is, for \{ \texttt{N} \} sets of tasks with two texts, to determine which of the two texts people in general would prefer. \\
\{ \texttt{end if} \}\\ \\
\{ \texttt{if stated preferences exist} \} 

- About Demographics

\{ \texttt{the annotator's demographics} \}

- About Reading Genre

\{ \texttt{the annotator's preferred reading genre} \}

- About Reading Frequency

\{ \texttt{the annotator's reading frequency} \}

- About Reading Motivation

\{ \texttt{the annotator's reading motivation} \}

- About Reading Preference

\{ \texttt{the annotator's preferred textual qualities} \}\\
\{ \texttt{end if} \}\\

\noindent \{ \texttt{if fewshot prompts exist} \} 

\{ \texttt{if inferring an annotator's preference} \}

\quad ===Consider the following example preference annotations from a user:===

\{ \texttt{else if inferring aggregated preference} \}

\quad ===Consider the following example preference annotations:===

\{ \texttt{end if} \}

\{ \texttt{for all fewshot prompts} \}

\quad Example \{ \texttt{i} \}-Text A:  

\quad \{ \texttt{text A} \} 
\\

\quad Example \{ \texttt{i} \}-Text B: 

\quad \{ \texttt{text B} \}

\quad \{ \texttt{if inferring an annotator's preference} \}

\quad \quad User's preference for Example \{i\}: \{ Text A, Text B, Unsure\}

\quad \{ \texttt{else if inferring aggregated preference} \}

\quad \quad Preference for Example \{i\}: \{ Text A, Text B, Unsure\}

\quad \{ \texttt{end if} \}

\{ \texttt{end for} \} \\
\{ \texttt{end if} \} \\ \\
\noindent \{ \texttt{if a synthesized profile exists} \}

===Below is the user persona description===

\{ synthesized profile \}\\
\{ \texttt{end if} \}\\

\noindent ===Your task is, for the following \{ \texttt{N} \} sets, to determine which text the user prefers: Text A or Text B.===\\
\{ \texttt{for all task pairs} \}

Set \{ \texttt{j} \}

- Text A:

\{ \texttt{text A} \}
\\

- Text B:

\{ \texttt{text B} \} \\
\{ \texttt{end for} \} \\ \\
=====\\
For each set, answer with "Text A" or "Text B" or "Unsure" if you cannot determine a preference. \\ \\
Do not provide any other information or reasoning, just the answer, in a list of answers. (e.g., ["Text A", "Text B", "Unsure"])
\end{framed}

Note that a prompt for each stated preference item is written as below.

\begin{framed}
\noindent Question: \{ \texttt{Question asked to the annotator} \}\\
Answer: \{ \texttt{Answer(s) selected by the annotator} \}
\end{framed}

\section{Interface for Evaluating RQ5 Analysis Pipeline}
\label{app:techeval_interface}

\begin{figure}
    \centering
    \includegraphics[width=0.478\textwidth]{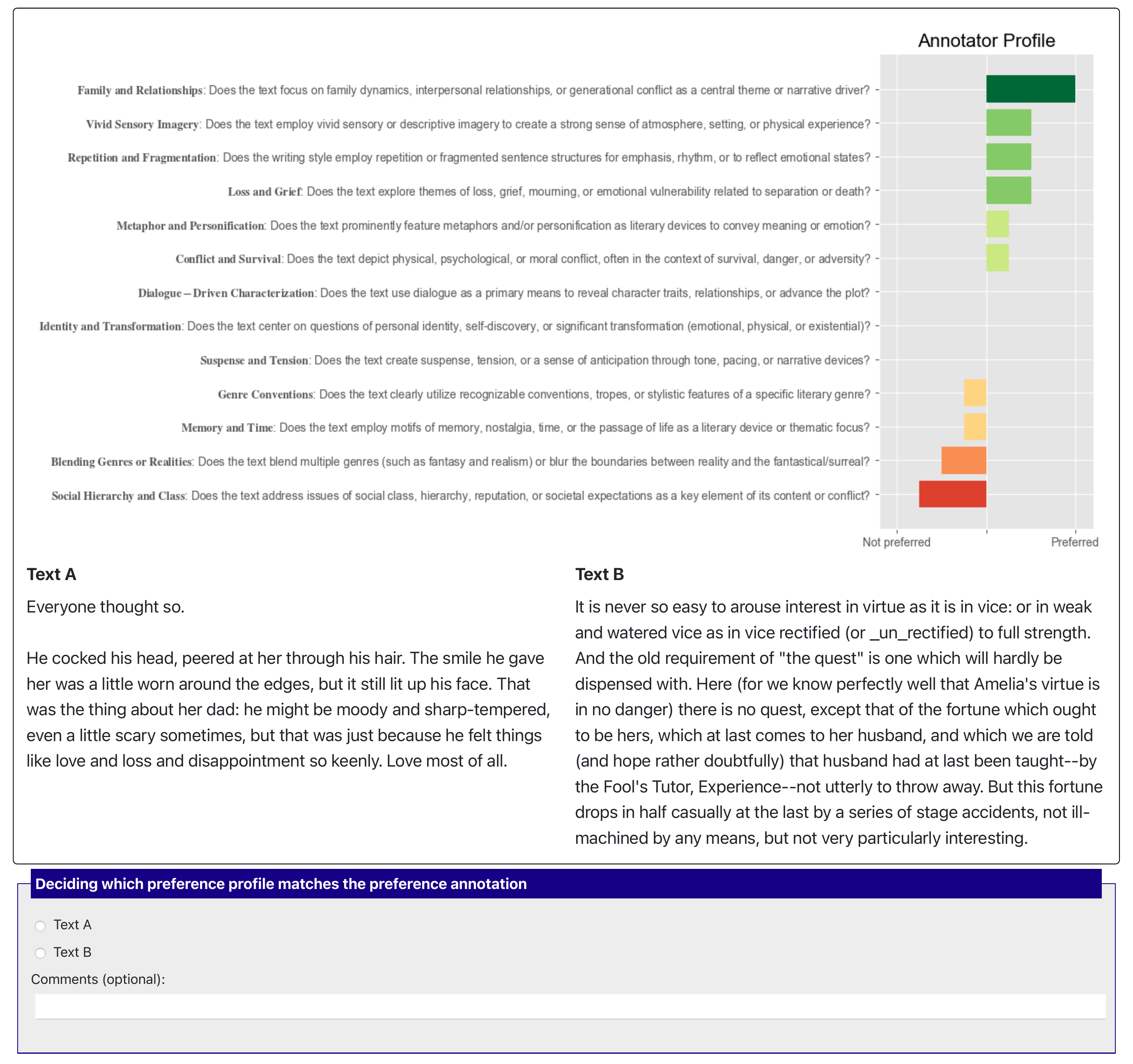}
    \caption{Interface for technical evaluation of RQ5 analysis pipeline.}
    \label{fig:techeval_interface}
    \Description{This figure presents an interface for technically evaluating the quality of preference vectors, by guessing which text should be preferred from a pair of texts given a preference vector. At the top of the interface, an "Annotator Profile" section displays a horizontal bar chart showing the annotator's preferences across the thirteen literary concepts identified in the study. The bars use a color gradient from dark green through light green to yellow, orange, and red, representing a spectrum from "Preferred" on the right to "Not preferred" on the left. Each concept is listed on the left with its defining question, while the bars extend rightward or leftward from a neutral center line to indicate preference strength. Below the profile visualization, the standard preference collection interface presents Text A and Text B side by side for comparison. The purple header below shows "Deciding which preference profile matches the preference annotation". The selection mechanism below offers radio buttons for Text A, Text B, and an optional comments field.}
\end{figure}

Figure~\ref{fig:techeval_interface} shows the interface used for data collection. Note that we represented the preference vector in a bar chart, with bars sorted in order from the most positive to the most negative.

\end{document}